\documentclass[english,letterpaper]{article}
\usepackage[T1]{fontenc}
\usepackage[latin9]{inputenc}
\usepackage{xcolor}
\usepackage{array}
\usepackage{textcomp}
\usepackage{multirow}
\usepackage{amsmath}
\usepackage{amsthm}
\usepackage{amssymb}
\usepackage{graphicx}

\makeatletter

\providecommand{\tabularnewline}{\\}

\usepackage{aaai21}  
\usepackage{times}  
\usepackage{helvet} 
\usepackage{courier}  
\usepackage[hyphens]{url}  
\usepackage{graphicx} 
\usepackage{natbib}  
\usepackage{caption} 
\usepackage{flushend}
\usepackage{microtype}

\@ifundefined{showcaptionsetup}{}{%
 \PassOptionsToPackage{caption=false}{subfig}}
\usepackage{subfig}
\makeatother

\usepackage{babel}
\begin{document}
\pdfinfo{ 
/Title (Semi-Supervised Learning with Variational Bayesian Inference and Maximum Uncertainty Regularization) 
/Author (Kien Do, Truyen Tran, Svetha Venkatesh) 
/TemplateVersion (2021.2) 
}

\setcounter{secnumdepth}{0}
\title{Semi-Supervised Learning with Variational Bayesian Inference and Maximum Uncertainty Regularization}
\author{Kien Do, Truyen Tran, Svetha Venkatesh\\}
\affiliations{
	Applied Artificial Intelligence Institute (A2I2), Deakin University, Geelong, Australia \\
	\{k.do, truyen.tran, svetha.venkatesh\}@deakin.edu.au
}
\maketitle
\begin{abstract}
We propose \emph{two} generic methods for improving semi-supervised
learning (SSL). The first integrates \emph{weight perturbation} (WP)
into existing ``consistency regularization'' (CR) based methods.
We implement WP by leveraging variational Bayesian inference (VBI).
The second method proposes a novel consistency loss called ``maximum
uncertainty regularization'' (MUR). While most consistency losses
act on perturbations in the vicinity of each data point, MUR actively
searches for \emph{``virtual''} points situated beyond this region
that cause the most uncertain class predictions. This allows MUR to
impose smoothness on a wider area in the input-output manifold. Our
experiments show clear improvements in classification errors of various
CR based methods when they are combined with VBI or MUR or both.

\end{abstract}
\global\long\def\Expect{\mathbb{E}}
\global\long\def\Real{\mathbb{R}}
\global\long\def\Data{\mathcal{D}}
\global\long\def\Loss{\mathcal{L}}
\global\long\def\Normal{\mathcal{N}}
\global\long\def\sg{\text{sg}}
\global\long\def\argmin#1{\underset{#1}{\text{argmin }}}
\global\long\def\argmax#1{\underset{#1}{\text{argmax }}}

\section{Introduction\label{sec:Introduction}}

Recent success in training deep neural networks is mainly attributed
to the availability of large, labeled datasets. However, annotating
large amounts of data is often expensive and time-consuming, and sometimes
requires specialized expertise (e.g., healthcare). Under these circumstances,
semi-supervised learning (SSL) has proven to be an effective means
of mitigating the need for labels by leveraging unlabeled data to
considerably improve performance. Among a wide range of approaches
to SSL \cite{van2019survey}, \emph{``consistency regularization''}
(CR) based methods are currently state-of-the-art \cite{bachman2014learning,sajjadi2016regularization,laine2016temporal,tarvainen2017mean,miyato2018virtual,verma2019interpolation,xie2019unsupervised,berthelot2019mixmatch,sohn2020fixmatch}.
These methods encourage neighbor samples to share labels by enforcing
consistent predictions for inputs under perturbations.

\begin{figure}
\begin{centering}
\includegraphics[height=0.19\textwidth]{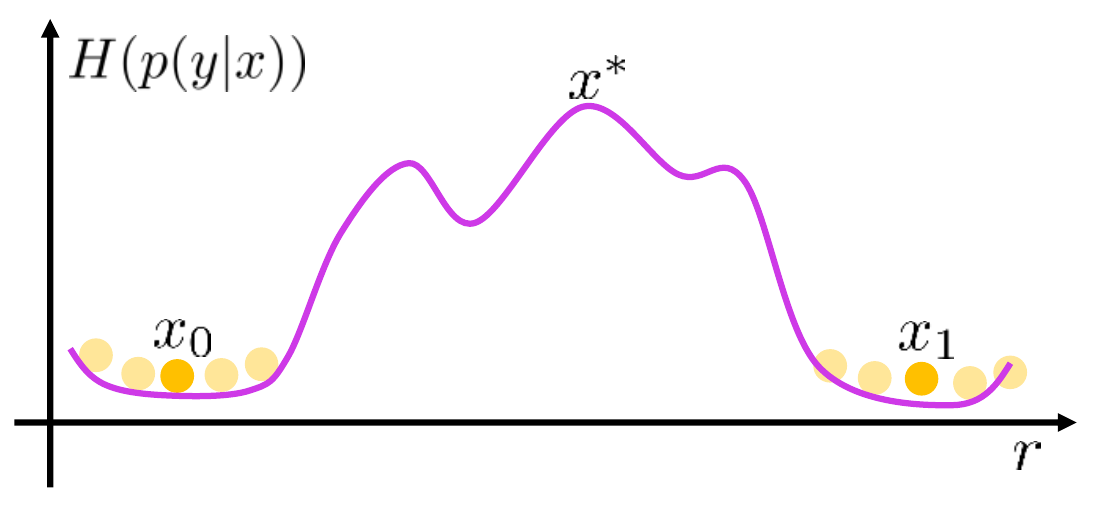}
\par\end{centering}
\caption{The most uncertain \emph{``virtual''} point $x^{*}$ usually lies
outside the vicinity of $x_{0}$ and is the most disruptive.\label{fig:MUR-idea}}
\end{figure}

Although the perturbations can be created in either the input/feature
space (data perturbation) or the weight space (weight perturbation),
existing CR based methods focus exclusively on the former and leave
the latter underexplored. Despite being related, \emph{weight perturbation}
(WP) is inherently different from data perturbation (DP) in the sense
that WP directly reflects different views of the classifier $f$ on
the original data distribution (e.g., $f_{w}(x)$ v.s. $f_{w'}(x)$)
while DP indirectly causes the classifier to adjust its view to adapt
to different data distributions (e.g., $f_{w}(x)$ v.s. $f_{w}(x')$).
Therefore, we hypothesize that the two types of perturbations are
complementary and could be combined to increase the classifier's robustness.
To implement WP, we treat the classifier's weights $w$ as random
variables and perform variational Bayesian inference (VBI) on $w$.
This approach has several advantages. First, perturbations of $w$
can be generated easily via drawing samples from an explicit variational
distribution $q_{\phi}(w|\Data)$. We also take advantage of the local
reparameterization trick and variational dropout (VD) \cite{kingma2015variational,molchanov2017variational}
to substantially reduce the sampling cost and variance of the gradients
w.r.t. the weight samples, making VBI scalable for deep neural networks.
Second, since VBI is a realization of the Minimum Description Length
(MDL) framework \cite{hinton93keeping,honkela2004variational}, a
classifier trained under VBI, in principle, often generalize better
than those trained in the standard way.

Standard DP methods (e.g., Gaussian noise, dropout) often generate
perturbations in the vicinity of each data point and ignore those
in the vacancy among data points, which means consistency losses equipped
with standard DPs can only train \emph{locally smooth} classifiers
that do not generalize well in general. To overcome this limitation,
we propose a novel consistency loss called \emph{``maximum uncertainty
regularization''} (MUR) of which the key component is finding for
each real data point $x_{0}$ a\emph{ }neighbor \emph{``virtual''}
point $x^{*}$ that has the most uncertain class prediction given
by the classifier $f$. During training, $f$ becomes increasingly
confident about its predictions of the real data points and their
neighborhood (otherwise, the training cannot converge). Thus, by choosing
$x^{*}$ with the above properties, we can guarantee, with high probability,
that i) $x^{*}$ is situated \emph{outside the vicinity} of $x_{0}$,
and ii) $x^{*}$ causes the biggest disruption to the classifier's
predictions (Fig.~\ref{fig:MUR-idea}). This observation suggests
that MUR enforces smoothness on a wider and rougher area in the input
space than conventional consistency losses, hence, making the classifier
generalize better. As MUR operates in the input space not the weight
space, it is complementary to WP and in some cases, both can be used
together.

In our experiments, we show that when strong data augmentation is
not available, WP and MUR significantly boost the performance of existing
CR based methods on various benchmark datasets.

\section{Preliminaries\label{sec:Preliminary}}

\subsection{Consistency Regularization based Methods for Semi-supervised Learning}

We briefly present two representative CR based methods namely $\Pi$-model
\cite{laine2016temporal} and Mean Teacher \cite{tarvainen2017mean}.
Other methods are discussed in the related work.

In $\Pi$-model, a classifier $f$ has \emph{deterministic} weights
$\theta$ but contains random input/feature perturbation layers such
as binary dropout \cite{srivastava2014dropout} and/or additive Gaussian
noise. Thus, if we pass an input sample $x$ through $f_{\theta}$
twice, we will get two different distributions $p(y|x,\theta)$ and
$p(y|x',\theta)$ where $x'$ is a perturbation of $x$ in the input/feature
space. The loss of $\Pi$-model is given by:
\begin{align}
\Loss_{\Pi}(\theta)=\  & \Expect_{(x_{l},y_{l})\sim\Data_{l}}\left[-\log p(y_{l}|x_{l},\theta)\right]+\nonumber \\
 & \lambda(t)\Expect_{x\sim\Data}\left[\frac{1}{K}\sum_{k=1}^{K}\left(p(k|x,\theta)-p(k|x',\theta_{\text{sg}})\right)^{2}\right]\label{eq:Pi_loss}\\
=\  & \Loss_{\text{xent,}l}(\theta)+\lambda(t)\Loss_{\Pi,\text{cons}}(\theta,\theta_{\text{sg}})\label{eq:Pi_loss_compact}
\end{align}
where $\Data_{l}$, $\Data_{u}$ denote the disjoint labeled and unlabeled
training datasets; $\Data=\Data_{l}\cup\Data_{u}$; $K$ is the number
of classes; $\lambda(t)$ is a ``ramp'' function which depends on
the training step $t$; $\theta_{\sg}$ denotes $\theta$ with no
gradient update; $\Loss_{\text{xent}}(\theta)$ is the cross-entropy
loss on labeled samples and $\Loss_{\Pi,\text{cons}}(\theta)$ is
the consistency loss on both labeled and unlabeled samples.

Mean Teacher (MT), on the other hand, introduces another network $f_{\bar{\theta}}$
called \emph{``mean teacher''} whose weights $\bar{\theta}$ are
the exponential moving averages (EMA) of $\theta$ across training
steps: $\bar{\theta}_{t}=\alpha\bar{\theta}_{t-1}+(1-\alpha)\theta$
($\alpha\in[0,1]$ is a momentum). Based on $f_{\bar{\theta}}$, MT
defines a new consistency loss $\Loss_{\text{MT,cons}}(\theta,\bar{\theta})$
between the outputs of $f_{\theta}$ and $f_{\bar{\theta}}$, which
leads to the final loss:
\begin{align}
\Loss_{\text{MT}}(\theta)=\  & \Expect_{(x_{l},y_{l})\sim\Data_{l}}\left[-\log p(y_{l}|x_{l},\theta)\right]+\nonumber \\
 & \lambda(t)\Expect_{x\sim\Data}\left[\frac{1}{K}\sum_{k=1}^{K}\left(p(k|x,\theta)-p(k|x',\bar{\theta})\right)^{2}\right]\label{eq:MT_loss}\\
=\  & \Loss_{\text{xent,}l}(\theta)+\lambda(t)\Loss_{\text{MT},\text{cons}}(\theta,\bar{\theta})\label{eq:MT_loss_compact}
\end{align}

\subsection{Variational Bayesian Inference and Variational Dropout\label{subsec:Variational-Bayesian-Inference}}

In Bayesian learning, we assume there is a prior distribution of weights,
denoted by $p(w)$. After observing the training dataset $\Data$,
we update our belief about $w$ as $p(w|\Data)=\frac{p(w)p(\Data|w)}{p(\Data)}=\frac{p(w)p(\Data|w)}{\int_{w}p(w)p(\Data|w)}$.
Since $p(w|\Data)$ is generally intractable to compute, we approximate
$p(w|\Data)$ with a variational distribution $q_{\phi}(w)$:
\begin{align}
 & \min_{\phi}D_{KL}\left(q_{\phi}(w)\|p(w|\Data)\right)\nonumber \\
= & \min_{\phi}\Expect_{w\sim q_{\phi}(w)}\left[\log\frac{q_{\phi}(w)}{p(w)p(\Data|w)}\right]\nonumber \\
= & \min_{\phi}\Expect_{w\sim q_{\phi}(w)}\left[-\log p(\Data|w)\right]+D_{KL}\left(q_{\phi}(w)\|p(w)\right)\label{eq:ELBO_w_1}
\end{align}
For discriminative classification tasks with an i.i.d. assumption
of data, Eq.~\ref{eq:ELBO_w_1} is equivalent to:
\begin{align}
\min_{\phi}\  & \Expect_{w\sim q_{\phi}(w)}\left[\Expect_{(x,y)\sim\Data}\left[-\log p(y|x,w)\right]\right]+\nonumber \\
 & \lambda D_{KL}\left(q_{\phi}(w)\|p(w)\right)\label{eq:ELBO_w_3}
\end{align}
where $x\in\Real^{M}$ and $y\in\{1,...,K\}$ denote an input sample
and its corresponding label, respectively; $\lambda$ is a balancing
coefficient, usually set to $\frac{1}{|\Data|}$. 

The loss function in Eq.~\ref{eq:ELBO_w_3} consists of two terms:
the expected negative data log-likelihood w.r.t. $q_{\phi}(w)$ and
the KLD between $q_{\phi}(w)$ and $p(w)$ which usually acts as a
regularization term on the model complexity. In order to minimize
this loss, we need to find a model that yields good classification
results yet still being as simple as possible. In fact, Eq.~\ref{eq:ELBO_w_3}
can also be viewed as the \emph{bits-back coding} objective under
the scope of the Minimum Description Length (MDL) principle \cite{graves2011practical,hinton93keeping,honkela2004variational}.

Generally, we could assume $q_{\phi}(w)$ to be a factorized Gaussian
distribution $q_{\phi}(w)=\Normal(w;\theta,\sigma^{2}\mathrm{I})$
and compute the gradient of the first term in Eq.~\ref{eq:ELBO_w_3}
w.r.t. $\theta$ and $\sigma$ using the reparameterization trick
\cite{kingma2013auto,rezende2014stochastic}. However, this approach
\cite{blundell2015weight} is computationally inefficient and has
high variance since it requires sampling of multiple weight components
for every data point. To handle this problem, Kingma et. al. \cite{kingma2015variational}
proposed the \emph{local reparameterization trick} and \emph{variational
dropout} (VD). Details about these techniques are given in the appendix.

\section{Our Approach\label{sec:Method}}

We describe how to integrate weight perturbation (WP) and maximum
uncertainty regularization (MUR) into CR based methods. Since WP is
realized via variational Bayesian inference (VBI), we will use VBI
in place of WP henceforth. We consider a general class of CR based
methods whose original objectives are of the form $\Loss_{\text{xent,}l}(\theta)+\lambda(t)\Loss_{\text{cons}}(\theta,\cdot)$,
denoted as $\mathcal{M}$. Any method $M\in\mathcal{M}$ can be combined
with VBI and MUR (denoted as $M$+VBI+MUR) by minimizing the following
loss:
\begin{align}
 & \Loss_{M\text{+VBI+MUR}}(\phi)=\nonumber \\
 & \Expect_{w\sim q_{\phi}(w)}\left[\Loss_{\text{xent,}l}(w)\right]+\lambda_{1}(t)\Expect_{w\sim q_{\phi}(w)}\left[\Loss_{M,\text{cons}}\left(w,\cdot\right)\right]+\nonumber \\
 & \lambda_{2}(t)D_{KL}\left(q_{\phi}(w)\|p(w)\right)+\lambda_{3}(t)\Expect_{w\sim q_{\phi}(w)}\left[\Loss_{\text{MUR}}(w)\right]\label{eq:M_VBI_MUR_generic}
\end{align}
where $q_{\phi}(w)=\Normal\left(w;\theta,\sigma^{2}\mathrm{I}\right)$
is the variational distribution of the classifier's weights $w$;
$\Loss_{\text{MUR}}$ is the MUR loss defined below; $\lambda_{1}(t)$,
$\lambda_{2}(t)$, $\lambda_{3}(t)$ are different ``ramp'' functions. 

Since $\Pi$-model and Mean Teacher are specific instances of $\mathcal{M}$,
we can easily derive the losses of $\Pi$+VBI+MUR and MT+VBI+MUR from
Eq.~\ref{eq:M_VBI_MUR_generic} by replacing $\Loss_{M,\text{cons}}(w,\cdot)$
with $\Loss_{\Pi,\text{cons}}(w,\theta_{\text{sg}})$ (Eq.~\ref{eq:Pi_loss_compact})
and $\Loss_{\text{MT,cons}}(w,\bar{\theta})$ (Eq.~\ref{eq:MT_loss_compact}),
respectively. For other CR based methods $M'$ having additional loss
terms $\Loss'(\theta)$ apart from $\Loss_{\text{xent},l}(\theta)$
and $\Loss_{\text{cons}}(\theta,\cdot)$, we can still construct the
loss of $M'$+VBI+MUR by simply adding $\Expect_{q_{\phi}(w)}\left[\Loss'(w)\right]$
to the RHS of Eq.~\ref{eq:M_VBI_MUR_generic}.

$M$+VBI+MUR has two special cases which are $M$+VBI and $M$+MUR.
The loss of $M$+VBI ($\Loss_{M\text{+VBI}}$) is similar to $\Loss_{M\text{+VBI+MUR}}$
but with the last term on the RHS discarded (e.g., by setting $\lambda_{3}(t)=0\ \forall t$).
On the other hand, by removing the third term as well as the expectation
w.r.t. $q_{\phi}(w)$ in all the remaining terms on the RHS of Eq.~\ref{eq:M_VBI_MUR_generic},
we obtain the loss of $M$+MUR ($\Loss_{M\text{+MUR}}$).

It is important to note that the second and the last terms on the
RHS of Eq.~\ref{eq:M_VBI_MUR_generic} are novel and have never been
used for SSL. While the second term shares some similarity with ensemble
learning in which different views of a classifier are combined to
obtain a robust prediction for a particular training example, the
last term is more related to multi-view learning \cite{qiao2018deep}
as different classifiers are applied to different views of data. However,
compared to ensemble learning and multi-view learning, our approach
is much more efficient since we can have almost infinite numbers of
views without training multiple classifiers.

\subsection{Weight Perturbation via Variational Bayesian Inference}

In Eq.~\ref{eq:M_VBI_MUR_generic}, we perturb the classifier's weights
$w$ by drawing random samples from $q_{\phi}(w)$. Minimizing $\Loss_{\text{xent,}l}(w)$,
$\Loss_{M,\text{cons}}(w,\cdot)$, and $\Loss_{\text{MUR}}(w)$ makes
the classifier robust against different weight perturbations while
minimizing $D_{KL}\left(q_{\phi}(w)\|p(w)\right)$ prevents the classifier
from being too complex. Both improve the classifier's generalizability.

We can see that the first and the third terms on the RHS of Eq.~\ref{eq:M_VBI_MUR_generic}
form a VBI objective similar to the one in Eq.~\ref{eq:ELBO_w_3}
but with the negative log-likelihood computed on labeled data only.
Due to the scarcity of labels in SSL, it seems reasonable to take
into account of the unlabeled data to model $q_{\phi}(w)$ better
by adding $\Expect_{w\sim q_{\phi}(w)}\left[\Expect_{x\sim\Data}\left[-\log p(x|w)\right]\right]$
to Eq.~\ref{eq:M_VBI_MUR_generic}. However, there are some difficulties:
i) we need to create an additional model for $p(x)$ that \emph{shares
weights} with the default classifier, and ii) the impact of modeling
both $p(x|w)$ and $p(y|x,w)$ using the same $w$ on the classifier's
generalizability is unclear. We observe empirically that the loss
in Eq.~\ref{eq:M_VBI_MUR_generic} produces good results, thus, we
leave modeling $p(x|w)$ for future work with a note that the work
by Grathwohl et. al. \cite{grathwohl2019your} may provide a good
starting point. To implement VBI, we adopt the variational dropout
(VD) technique from \cite{molchanov2017variational}. Our justification
for this is presented in the appendix.

\subsection{Maximum Uncertainty Regularization\label{subsec:Most-Uncertainty-Regularization}}

Some CR based methods enforce smoothness on the vicinity of training
data points by using standard data perturbation (DP) techniques (e.g.,
Gaussian noise, dropout). However, there are points in the input-output
manifold unreachable by standard DPs. These \emph{``virtual''} points
usually lie beyond the local area of real data points and prevent
a smooth transition of the class prediction from a data point to another.
We argue that if we can find such ``virtual'' points and force their
class predictions to be similar to those of nearby data points, we
will learn a smoother classifier that generalizes better. We do this
by introducing a novel consistency loss called \emph{maximum uncertainty
regularization} (MUR) $\Loss_{\text{MUR}}$. In case the classifier's
weights $\theta$ are \emph{deterministic}, $\Loss_{\text{MUR}}$
is given by:
\begin{equation}
\Loss_{\text{MUR}}(\theta)=\Expect_{x_{0}\sim\Data}\left[\frac{1}{K}\sum_{k=1}^{K}\left(p(k|x^{*},\theta)-p(k|x_{0},\theta_{\sg})\right)^{2}\right]\label{eq:MUR_loss}
\end{equation}
where $x^{*}$ is mathematically defined as follows:
\begin{equation}
x^{*}=\argmax xH(p(y|x))\ \ \ \text{s.t.}\ \ \ \left\Vert x-x_{0}\right\Vert _{2}\leq r\label{eq:x_star}
\end{equation}
where $H(\cdot)$ is the Shannon entropy, $r\in\Real^{+}$ is the
largest distance between $x^{*}$ and $x_{0}$. In general, it is
hard to compute $x^{*}$ exactly because the objective in Eq.~\ref{eq:x_star}
usually has many local optima. However, we can approximate $x^{*}$
by optimizing a linear approximation of $H(p(y|x))$ instead. In this
case, the original optimization problem becomes \emph{convex minimization}
and it has a unique solution which is:
\begin{equation}
x^{*}\approx\tilde{x}^{*}=x_{0}+r\frac{g_{0}}{\left\Vert g_{0}\right\Vert _{2}}\label{eq:x_star_approx}
\end{equation}
where $g_{0}=\frac{\partial H(p(y|x))}{\partial x}\big|_{x=x_{0}}$
is the gradient of $H(p(y|x))$ at $x=x_{0}$. Its derivation is presented
in the appendix.

\subsubsection{Iterative approximations of $x^{*}$}

Linearly approximating $H(p(y|x))$ may cause some information loss.
We can avoid that by optimizing Eq.~\ref{eq:x_star} directly via
\emph{projected gradient ascent} (details in the appendix). Alternatively,
\emph{vanilla gradient ascent} update based on the Lagrangian relaxation
of Eq.~\ref{eq:x_star} can be done via maximizing:
\begin{equation}
\mathcal{F}(x)=H(p(y|x))-\lambda^{*}(x)\left(\left\Vert x-x_{0}\right\Vert _{2}-r\right)\label{eq:Iterative_Lagrange}
\end{equation}
where $\lambda^{*}(x)=\frac{\left\Vert x-x_{0}\right\Vert _{2}\left\Vert g_{0}\right\Vert _{2}}{r}$.
Insight on $\lambda^{*}(x)$ is given in the appendix.

\subsubsection{Connection to Adversarial Learning}

Adversarial learning (AL) \cite{szegedy2013intriguing} aims to build
a system robust against various types of adversarial attacks. Madry
et. al. \cite{madry2017towards} have shown that these methods attempt
to solve the following saddle point problem:
\begin{equation}
\min_{\theta}\Expect_{(x,y)\sim\Data}\left[\sup_{\epsilon\in\mathcal{S}}L\left(f_{\theta}\left(x+\epsilon\right),y\right)\right]\label{eq:Adversarial_Learning}
\end{equation}
where $L$ is a loss function (e.g., the cross-entropy), $\mathcal{S}$
is the support set of the adversarial noise $\epsilon$. For example,
in case of Fast Gradient Sign Method \cite{goodfellow2014explaining},
$\mathcal{S}$ is defined as $\mathcal{S}=\left\{ \epsilon:\left\Vert \epsilon\right\Vert _{\infty}\leq r\right\} $.
Adversarial learning has also been shown by Sinha et. al. \cite{sinha2017certifying}
to be closely related to \emph{(distributionally) robust optimization}
\cite{farnia2016minimax,globerson2006nightmare} whose objective is
given by:
\begin{equation}
\min_{\theta}\sup_{p\in\mathcal{P}(\Data)}\Expect_{(x,y)\sim p}\left[L\left(f_{\theta}(x),y\right)\right]\label{eq:Robust_Optimization}
\end{equation}
where $\mathcal{P}(\Data)$ is a class of distributions derived from
the empirical data distribution $\Data$.

At the high level, MUR (Eqs.~\ref{eq:MUR_loss}, \ref{eq:x_star})
is similar to AL (Eq.~\ref{eq:Adversarial_Learning}) as both consist
of two optimization sub-problems: an \emph{inner maximization w.r.t.
the data} and an \emph{outer minimization w.r.t. the parameters}.
However, when looking closer, there are some differences between MUR
and AL: In MUR, the two sub-problems optimize two distinct objectives
(the consistency loss and the conditional entropy) while in AL, the
two sub-problems share the same objective. Moreover, since MUR's objectives
do not use label information, MUR is applicable to SSL while AL is
not. 

Compared to virtual adversarial training (VAT) \cite{miyato2018virtual},
MUR is different in how $x^{*}$ is chosen. VAT defines $x^{*}$ to
be a point in the local neighborhood of $x_{0}$ whose output $p(y|x^{*})$
is the most different from $p(y|x_{0})$. It means that $x^{*}$ can
have very low $H(p(y|x^{*}))$ as long as its corresponding pseudo
class is different from the (pseudo) class of $x_{0}$. MUR, by contrast,
always looks for $x^{*}$ with the highest $H(p(y|x^{*}))$ regardless
of the (pseudo) class of $x_{0}$. Inspired by VAT and MUR, we propose
a new CR based method called\emph{ maximum uncertainty training} (MUT)
with the loss function defined as:
\[
\Loss_{\text{MUT}}(\theta)=\Loss_{\text{xent},l}(\theta)+\lambda(t)\Loss_{\text{MUR}}(\theta)
\]

MUT can be seen as a special case of $\Pi$+MUR in which the coefficient
of $\Loss_{\Pi\text{,cons}}(\theta,\theta_{\text{sg}})$ equals 0.
We can also view it as a variant of $\Pi$-model (Eq.~\ref{eq:Pi_loss_compact})
with $\Loss_{\Pi\text{,cons}}(\theta,\theta_{\text{sg}})$ replaced
by $\Loss_{\text{MUR}}(\theta)$. Note that for other CR based methods
like MT or ICT, their original consistency losses cannot be replaced
by $\Loss_{\text{MUR}}$ since these losses and $\Loss_{\text{MUR}}$
are inherently different. For example, in MT, $\Loss_{\text{MT,cons}}(\theta,\bar{\theta})$
involves both the student and teacher networks while $\Loss_{\text{MUR}}(\theta)$
only involves the student network.

\section{Experiments\label{sec:Experiments}}

We now show that using VBI (or VD in particular) and MUR leads to
significant improvements in performance and generalization of CR based
methods that do not use strong data augmentation. For methods that
use strong data augmentation (e.g., FixMatch \cite{sohn2020fixmatch}),
results are discussed in the appendix. We evaluate our approaches
on three standard benchmark datasets: SVHN, CIFAR-10 and CIFAR-100.
Details about the datasets, data preprocessing scheme, the classifier's
architecture and settings, and the training hyperparameters are all
provided in the appendix.

\subsection{Classification results on SVHN, CIFAR-10 and CIFAR-100}

In Tables \ref{tab:CIFAR10_CIFAR100_results} and \ref{tab:SVHN_results},
we compare the classification errors of state-of-the-art CR based
methods with/without using VD and MUR on SVHN, CIFAR-10, and CIFAR-100.
Results of the baselines are taken from existing literature. We provide
results from our own implementations of some baselines when necessary.
Each setting of our models is run 3 times.

\paragraph{SVHN}

When there are 500 and 1000 labeled samples, combining $\Pi$ with
MUR reduces the error by about 1-2\% compared to the plain one. In
case VD is used instead of MUR, the error reduction is about 0.5-0.9\%.
It suggests that MUR is more helpful for $\Pi$ than VD. On the other
hand, when the base model is MT, using VD leads to bigger improvements
than using MUR. Interestingly, using both VD and MUR for MT boosts
the performance even further. By contrast, using both VD and MUR for
$\Pi$ leads to higher error with larger variance compared to using
individual methods. We think the main cause is the inherent instability
of $\Pi$ as this model does not use weight averaging for prediction
like MT. Thus, too much randomness from both VD and MUR can be harmful
for $\Pi$.

\begin{table*}
\begin{centering}
\begin{tabular}{l|ccc|c}
\hline 
\multirow{2}{*}{Model} & \multicolumn{3}{c|}{CIFAR-10} & CIFAR-100\tabularnewline
 & 1000 & 2000 & 4000 & 10000\tabularnewline
\hline 
\hline 
$\Pi$$^{\heartsuit}$ & 31.65 $\pm$1.20 & 17.57$\pm$0.44 & 12.36$\pm$0.31 & 39.19$\pm$0.54\tabularnewline
$\Pi$ + FSWA$^{\diamondsuit}$  & 17.23$\pm$0.34 & 12.61$\pm$0.18 & 10.07$\pm$0.27 & 34.25$\pm$0.16\tabularnewline
\hline 
TempEns + SNTG$^{\clubsuit}$ & 18.14$\pm$0.52 & 13.64$\pm$0.32 & 10.93$\pm$0.14 & -\tabularnewline
\hline 
VAT$^{\spadesuit}$ & - & - & 10.55$\pm$0.05 & -\tabularnewline
\hline 
MT$^{\heartsuit}$ & 21.55$\pm$1.48 & 15.73$\pm$0.31 & 12.31$\pm$0.28 & -\tabularnewline
MT$^{\diamondsuit}$ & 18.78$\pm$0.31 & 14.43$\pm$0.20 & 11.41$\pm$0.27 & 35.96$\pm$0.77\tabularnewline
MT + FSWA$^{\diamondsuit}$ & 15.58$\pm$0.12 & 11.02$\pm$0.23 & 9.05$\pm$0.21 & 33.62$\pm$0.54\tabularnewline
\textcolor{purple}{MT} & \textcolor{purple}{19.63$\pm$0.33} & \textcolor{purple}{15.07$\pm$0.10} & \textcolor{purple}{11.65$\pm$0.09} & \textcolor{purple}{37.65$\pm$0.25}\tabularnewline
\textcolor{purple}{MT + VD} & \textcolor{purple}{16.35$\pm$0.18} & \textcolor{purple}{12.51$\pm$0.43} & \textcolor{purple}{9.62$\pm$0.13} & \textcolor{purple}{35.47$\pm$0.21}\tabularnewline
\textcolor{purple}{MT + MUR} & \textcolor{purple}{17.96$\pm$0.32} & \textcolor{purple}{12.23$\pm$0.21} & \textcolor{purple}{10.16$\pm$0.04} & \textcolor{purple}{35.93$\pm$0.32}\tabularnewline
\textcolor{purple}{MT + VD + MUR} & \textbf{\textcolor{purple}{15.47$\pm$0.13}} & \textbf{\textcolor{purple}{10.57$\pm$0.28}} & \textbf{\textcolor{purple}{8.54$\pm$0.20}} & \textcolor{purple}{35.24$\pm$0.06}\tabularnewline
\hline 
ICT$^{\spadesuit}$  & 15.48$\pm$0.78 & 9.26$\pm$0.09 & 7.92$\pm$0.02 & -\tabularnewline
\textcolor{purple}{ICT} & \textcolor{purple}{14.15$\pm$0.16} & \textcolor{purple}{11.56$\pm$0.07} & \textcolor{purple}{9.18$\pm$0.03} & \textcolor{purple}{35.67$\pm$0.07}\tabularnewline
\textcolor{purple}{ICT + VD} & \textbf{\textcolor{purple}{10.13$\pm$0.21}} & \textcolor{purple}{8.83$\pm$0.15} & \textbf{\textcolor{purple}{7.48$\pm$0.11}} & \textcolor{purple}{34.12$\pm$0.16}\tabularnewline
\textcolor{purple}{ICT + MUR} & \textcolor{purple}{13.54$\pm$0.23} & \textcolor{purple}{10.49$\pm$0.07} & \textcolor{purple}{8.55$\pm$0.06} & \textcolor{purple}{34.91$\pm$0.20}\tabularnewline
\textcolor{purple}{ICT + VD + MUR} & \textcolor{purple}{10.37$\pm$0.25} & \textbf{\textcolor{purple}{8.79$\pm$0.16}} & \textcolor{purple}{7.55$\pm$0.14} & \textbf{\textcolor{purple}{33.21$\pm$0.24}}\tabularnewline
\hline 
Co-train (8 views)$^{\dagger}$ & - & - & 8.35$\pm$0.06 & -\tabularnewline
\hline 
\end{tabular}
\par\end{centering}
\smallskip{}

\caption{Classification errors on CIFAR-10 and CIFAR-100. $^{\heartsuit}$:
\cite{tarvainen2017mean}, $^{\diamondsuit}$: \cite{athiwaratkun2018there},
$^{\clubsuit}$: \cite{luo2018smooth}, $^{\spadesuit}$: \cite{verma2019interpolation},
$^{\dagger}$: \cite{qiao2018deep}. Our implementations are highlighted
in purple.\label{tab:CIFAR10_CIFAR100_results}}
\end{table*}

When the number of labels is 250, our implementations of $\Pi$ and
MT yield much poorer results than the original models. However, we
note that the same problem can also be found in \cite{berthelot2019mixmatch}
(Table 6) and \cite{sohn2020fixmatch} (Table 2). Therefore, to ensure
fair comparison, we only consider the results of our implementations.
While using VD still improves the performances of $\Pi$ and MT, using
MUR hurts the performances of these models. A possible reason is that
with too few labeled examples, the classifier is unable to learn correct
class-intrinsic features (unless strong data augmentation is given),
hence, the gradient of $H(p(y|x))$ w.r.t. $x$ (Eq.~\ref{eq:x_star})
may point to wrong directions.

\paragraph{CIFAR-10/CIFAR-100}

We observe the same pattern for MT on CIFAR-10 and CIFAR-100 as on
SVHN: Using VD+MUR leads to much better results than using either
VD or MUR. Specifically on CIFAR-10, VD+MUR decreases the errors of
MT by about 3-4.5\% while for VD and MUR, the amounts of error reduction
are 2-2.7\% and 1.5-2.8\%, respectively. Compared to MT+FSWA \cite{athiwaratkun2018there},
our MT+VD+MUR achieves slightly better results on CIFAR-10 but perform
worse on CIFAR-100. The reason could be that they use better settings
for CIFAR-100 than ours, which is reflected in the lower error of
their MT compared to our reimplemented MT. However, we want to note
that FSWA only provides MT with advanced learning rate scheduling
\cite{loshchilov2016sgdr} and postprocessing \cite{izmailov2018averaging}
but does not change the objective of MT like VD or MUR. It means one
can easily combine MT+VD+MUR with FSWA to further improve the results.

In case of ICT, using VD leads to impressive decreases of error by
1.7-4\% on CIFAR-10 and by 1.5\% on CIFAR-100. Meanwhile, MUR only
improves the results slightly, by about 0.6-1\% on CIFAR-10 and by
0.7\% on CIFAR-100. The performance of ICT+VD+MUR is also just comparable
to that of ICT+VD. A possible reason is that because ICT enforces
smoothness on points interpolated between pairs of random real data
points which, to some extent, are similar to the ``virtual'' points
in MUR. Thus, the regularization effect of ICT may overlap that of
MUR.

\begin{table}
\begin{centering}
{\small{}}%
\begin{tabular}{l|ccc}
\hline 
{\small{}Model} & {\small{}250} & {\small{}500} & {\small{}1000}\tabularnewline
\hline 
\hline 
{\small{}$\Pi$$^{\heartsuit}$} & {\small{}9.69$\pm$0.92} & {\small{}6.65$\pm$0.53} & {\small{}4.82$\pm$0.17}\tabularnewline
{\small{}$\Pi$ + SNTG$^{\diamondsuit}$} & {\small{}5.07$\pm$0.25} & {\small{}4.52$\pm$0.30} & {\small{}3.82$\pm$0.25}\tabularnewline
\textcolor{purple}{\small{}$\Pi$} & \textcolor{purple}{\small{}13.37$\pm$0.97} & \textcolor{purple}{\small{}7.25$\pm$0.36} & \textcolor{purple}{\small{}5.18$\pm$0.13}\tabularnewline
\textcolor{purple}{\small{}$\Pi$ + VD} & \textcolor{purple}{\small{}12.98$\pm$0.84} & \textcolor{purple}{\small{}6.38$\pm$0.42} & \textcolor{purple}{\small{}4.65$\pm$0.23}\tabularnewline
\textcolor{purple}{\small{}$\text{\ensuremath{\Pi}}$ + MUR} & \textcolor{purple}{\small{}15.04$\pm$0.75} & \textcolor{purple}{\small{}5.43$\pm$0.27} & \textcolor{purple}{\small{}4.15$\pm$0.10}\tabularnewline
\textcolor{purple}{\small{}$\text{\ensuremath{\Pi}}$ + VD + MUR} & \textcolor{purple}{\small{}16.63$\pm$1.22} & \textcolor{purple}{\small{}6.57$\pm$0.73} & \textcolor{purple}{\small{}4.72$\pm$0.48}\tabularnewline
\hline 
{\small{}VAT$^{\clubsuit}$} & {\small{}-} & {\small{}-} & {\small{}3.86$\pm$0.11}\tabularnewline
\hline 
{\small{}MT$^{\heartsuit}$} & {\small{}4.53$\pm$0.50} & {\small{}4.18$\pm$0.27} & {\small{}3.95$\pm$0.19}\tabularnewline
{\small{}MT + SNTG$^{\diamondsuit}$} & \textbf{\small{}4.29$\pm$0.43} & {\small{}3.99$\pm$0.24} & {\small{}3.86$\pm$0.27}\tabularnewline
\textcolor{purple}{\small{}MT} & \textcolor{purple}{\small{}5.57$\pm$1.52} & \textcolor{purple}{\small{}3.86$\pm$0.15} & \textcolor{purple}{\small{}3.72$\pm$0.10}\tabularnewline
\textcolor{purple}{\small{}MT + VD} & \textcolor{purple}{\small{}5.26$\pm$1.73} & \textcolor{purple}{\small{}3.39$\pm$0.10} & \textcolor{purple}{\small{}3.28$\pm$0.08}\tabularnewline
\textcolor{purple}{\small{}MT + MUR} & \textcolor{purple}{\small{}6.45$\pm$1.29} & \textcolor{purple}{\small{}3.66$\pm$0.07} & \textcolor{purple}{\small{}3.48$\pm$0.04}\tabularnewline
\textcolor{purple}{\small{}MT + VD + MUR} & \textcolor{purple}{\small{}6.82$\pm$2.01} & \textbf{\textcolor{purple}{\small{}3.21$\pm$0.13}} & \textbf{\textcolor{purple}{\small{}3.16$\pm$0.07}}\tabularnewline
\hline 
{\small{}ICT$^{\clubsuit}$} & {\small{}4.78$\pm$0.68} & {\small{}4.23$\pm$0.15} & {\small{}3.89$\pm$0.04}\tabularnewline
\hline 
{\small{}Co-train (8 views)$^{\spadesuit}$} & {\small{}-} & {\small{}-} & {\small{}3.29$\pm$0.03}\tabularnewline
\hline 
\end{tabular}{\small\par}
\par\end{centering}
\smallskip{}

\caption{Classification errors on SVHN. $^{\heartsuit}$: \cite{tarvainen2017mean},
$^{\diamondsuit}$: \cite{luo2018smooth}, $^{\clubsuit}$: \cite{verma2019interpolation},
$^{\spadesuit}$: \cite{qiao2018deep}. Our implementations are highlighted
in purple.\label{tab:SVHN_results}}
\end{table}

\begin{table}
\begin{centering}
\begin{tabular}{c|c|c}
\hline 
\multirow{2}{*}{} & \multicolumn{2}{c}{Sensitivity}\tabularnewline
\cline{2-3} 
 & MT & ICT\tabularnewline
\hline 
\hline 
Default & 0.21$\pm$0.26 & 0.22$\pm$0.44\tabularnewline
\hline 
+VD & 0.19$\pm$0.30 & 0.20$\pm$0.46\tabularnewline
\hline 
+MUR & 0.12$\pm$0.16 & 0.16$\pm$0.32\tabularnewline
\hline 
+VD+MUR & 0.13$\pm$0.29 & 0.19$\pm$0.38\tabularnewline
\hline 
\end{tabular}
\par\end{centering}
\smallskip{}

\caption{Sensitivities of MT, ICT and their variants trained on CIFAR-10 with
1000 labels.\label{tab:Sensitivity}}
\end{table}

\subsection{Effects of VBI and MUR on sensitivity\label{subsec:Results-on-sensitivity}}

Besides accuracy, we should also examine \emph{sensitivity} since
this metric is closely related to generalization \cite{alain2014regularized,novak2018sensitivity,rifai2011contractive}.
The\emph{ sensitivity} of a classifier $f$ w.r.t. small changes of
a data point $x$ is measured as the Frobenius norm of the Jacobian
matrix of $f$ w.r.t. $x$:
\[
\text{Sensitivity}(x)=\left\Vert J(x)\right\Vert _{F}=\sqrt{\sum_{i,j}J_{i,j}^{2}(x)}
\]
where $J_{i,j}(x)=\frac{\partial{f(x)}_{i}}{\partial x_{j}}=\frac{\partial p(y=i|x)}{\partial x_{j}}$.
A low sensitivity value means that the local area surrounding $x$
is flat\footnote{It should not be confused with flat minima \cite{hochreiter1997flat}
in the weight space.} and $f$ is robust to small variations of $x$.

Compared to MT and ICT, the corresponding variants using VD and/or
MUR achieve lower sensitivity on average (Table~\ref{tab:Sensitivity})
and have more test data points with sensitivities close to 0 (Figs.~2,
3 in the appendix). These results empirically verify that VD and
MUR actually make the classifier smoother.

\subsection{Ablation Study}

\begin{figure*}
\begin{centering}
\subfloat[classification error\label{fig:err_bar_kld_coeff}]{\begin{centering}
\includegraphics[height=0.19\textwidth]{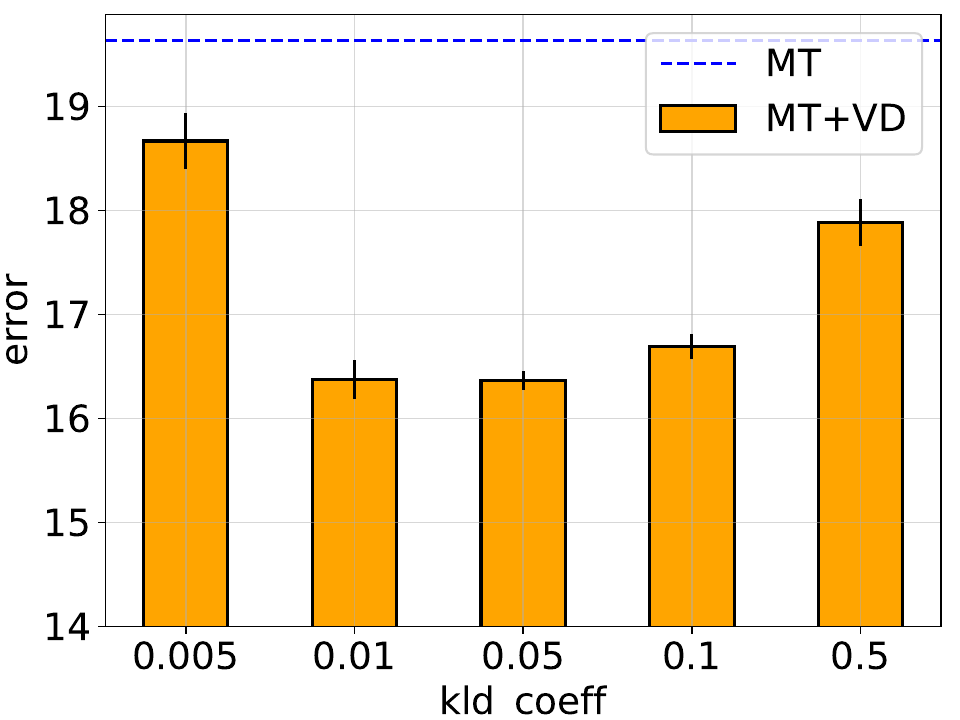}
\par\end{centering}
}\subfloat[sparsity curve\label{fig:sparsity_kld_coeff}]{\begin{centering}
\includegraphics[height=0.19\textwidth]{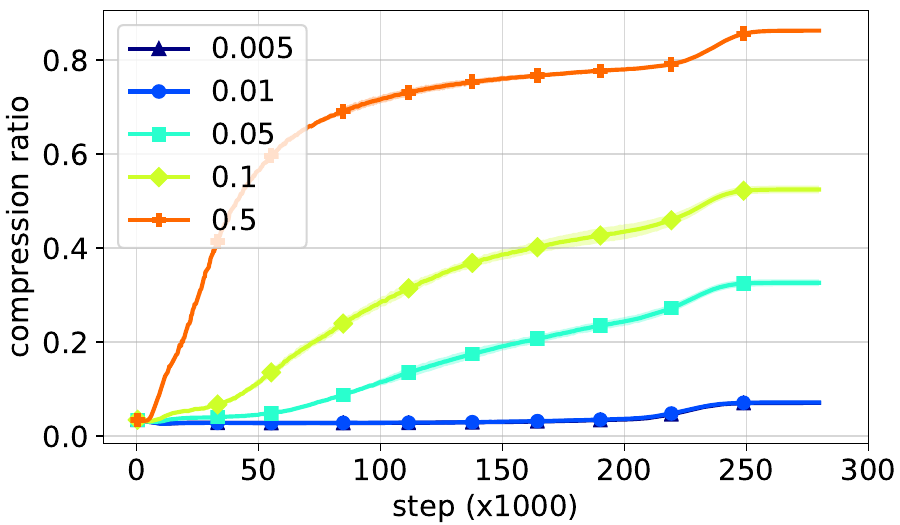}
\par\end{centering}
}\subfloat[classification error\label{fig:Error_as_fn_of_rad}]{\begin{centering}
\includegraphics[height=0.19\textwidth]{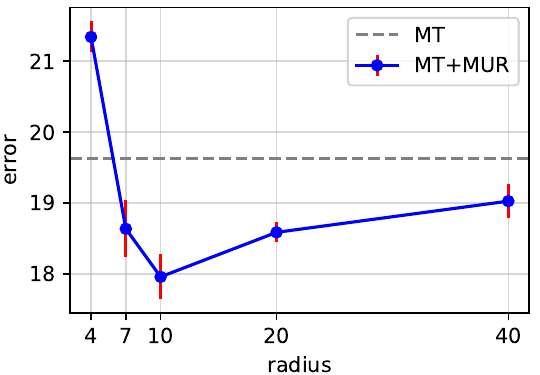}
\par\end{centering}
}
\par\end{centering}
\caption{(a), (b): Classification errors and weight sparsity curves of MT+VD
w.r.t. different coefficients of the KLD of weights. (c): Classification
errors of MT+MUR w.r.t. different values of the radius $r$. The dataset
is CIFAR-10 with 1000 labels.}
\end{figure*}

\begin{figure}
\begin{centering}
\includegraphics[height=0.19\textwidth]{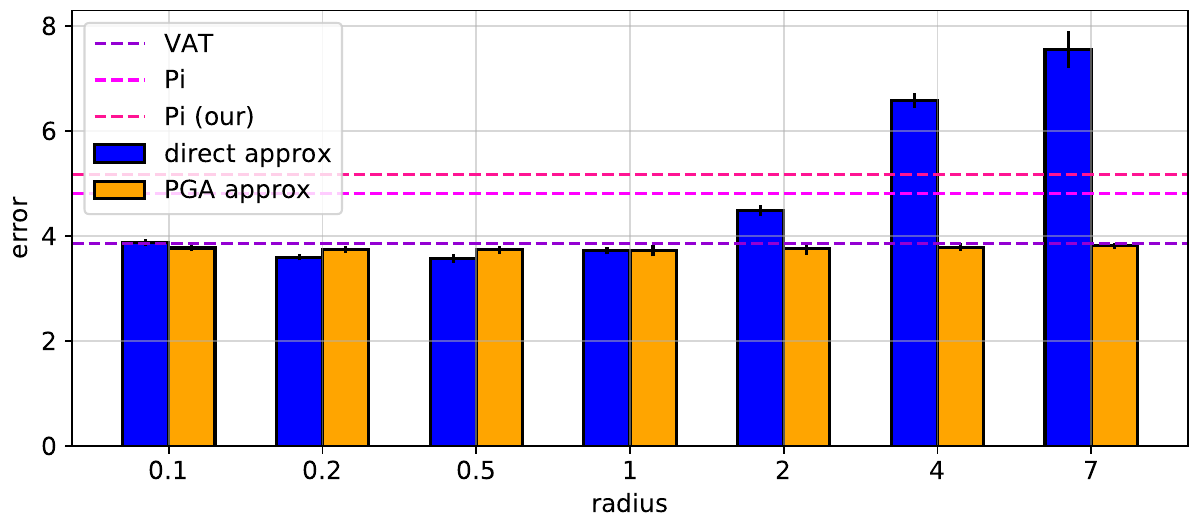}
\par\end{centering}
\caption{Classification errors of MUT w.r.t. different values of the radius
$r$. For the PGA update of $x^{*}$, lr=0.3 and \#steps=5. The dataset
is SVHN with 1000 labels.\label{fig:MUT_errors}}
\end{figure}

\paragraph{The coefficient of $D_{KL}\left(q_{\phi}(w)\|p(w)\right)$ in VBI}

In Fig.~\ref{fig:err_bar_kld_coeff}, we compare the errors of MT+VD
w.r.t. different values of the coefficient of $D_{KL}\left(q_{\phi}(w)\|p(w)\right)$\footnote{The ``coefficient'' in this context is referred to as the \emph{maximum}
value of $\lambda_{2}(t)$ in Eq.~\ref{eq:M_VBI_MUR_generic}}. Either too small or too large coefficients lead to inferior results
since they correspond to too little or too much regularization on
weights (Fig.~\ref{fig:sparsity_kld_coeff}). However, even in the
worst setting, MT+VD still outperforms MT. This again demonstrates
the clear advantage of VBI in improving the robustness of models.

\paragraph{The radius $r$ in MUR\label{par:The-radius-r}}

We now examine how the radius $r$ (Eq.~\ref{eq:x_star}) affects
performance. If $r$ is too small, it is hard to find an adequate
virtual point $x^{*}$ that the classifier $f$ is uncertain about.
Moreover, as $x^{*}$ is very close to $x_{0}$, minimizing $\Loss_{\text{MUR}}$
causes $f$ to focus too much on ensuring the local flatness around
$x_{0}$ instead of smoothing the area between $x_{0}$ and other
data points, exacerbating the problem. By contrast, if $r$ is too
big, $x^{*}$ is very different from $x_{0}$ and forcing consistency
between these points may be inappropriate. Fig.~\ref{fig:Error_as_fn_of_rad}
shows the error of MT+MUR on CIFAR-10 with 1000 labels as a function
of $r$ ($r\in\{4,7,10,20,40\}$), which reflects the intuition presented:
MT+MUR performs poorly when $r$ is too small (4) and worse than MT.
When $r$ is too big (20, 40), the results are also not good. The
optimal value of $r$ is 10. To make sure that this result is reasonable,
we visualize the virtual samples w.r.t. different values of $r$ in
the appendix.

\paragraph{Iterative approximations of $x^{*}$ in MUR}

\begin{figure*}
\begin{centering}
\subfloat[classification error\label{fig:iter_approx_err_bar}]{\begin{centering}
\includegraphics[height=0.19\textwidth]{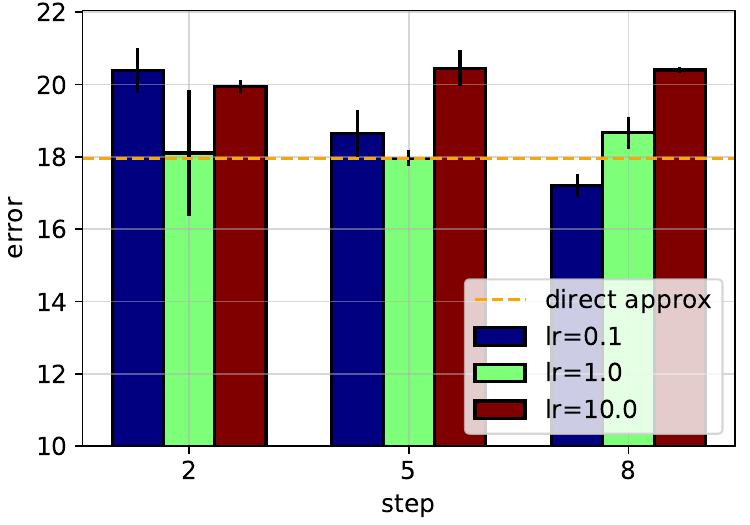}
\par\end{centering}
}\subfloat[$\left\Vert g_{0}\right\Vert _{2}$\label{fig:iter_approx_grad_norm_curve}]{\begin{centering}
\includegraphics[height=0.19\textwidth]{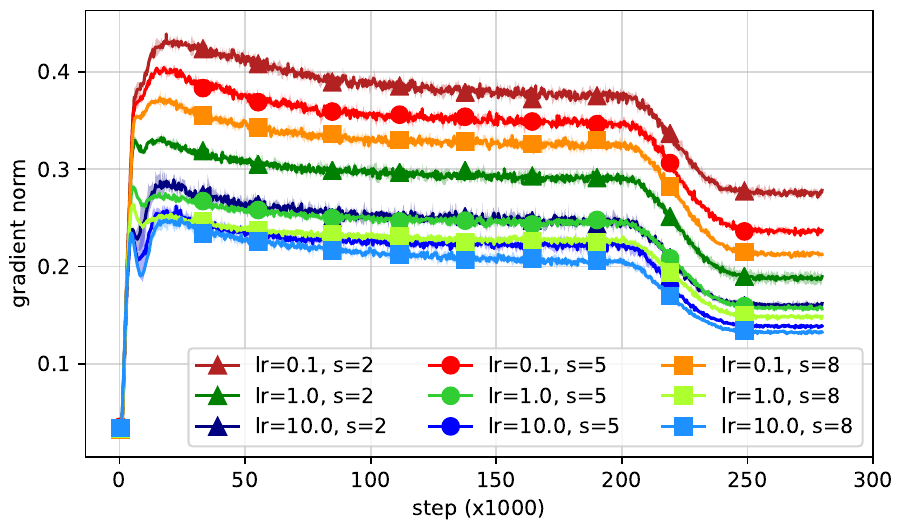}
\par\end{centering}
}\subfloat[classification error\label{fig:MT+RR}]{\begin{centering}
\includegraphics[height=0.19\textwidth]{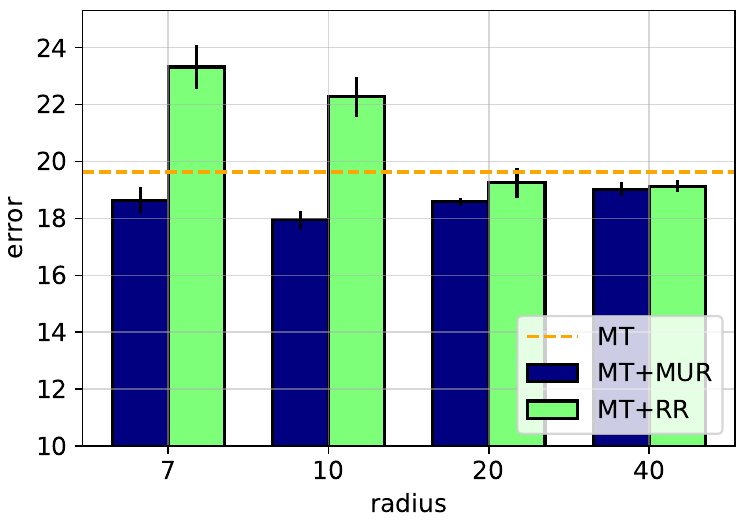}
\par\end{centering}
}
\par\end{centering}
\caption{(a), (b): Classification errors and gradient norm curves of MT+MUR
with the GA update of $x^{*}$ w.r.t. different learning rates and
numbers of steps. (c): Classification errors of MT+RR and MT+MUR w.r.t.
different values of the radius $r$. The dataset is CIFAR-10 with
1000 labels.\label{fig:iter_approx_GradAscent}}
\end{figure*}

We investigate the performance of MT+MUR when iterative approximations
of $x^{*}$ are used instead of the direct (linear) approximation
(Eq.~\ref{eq:x_star_approx}). We try both projected gradient ascent
(PGA) and vanilla gradient ascent (GA) updates with the learning rate
$\alpha$ varying in $\{0.1,1.0,10.0\}$ and the number of steps $s$
varying in $\{2,5,8\}$. We report results of the GA update in Fig.~\ref{fig:iter_approx_GradAscent}.
. Clearly, larger $\alpha$ and $s$ both lead to smaller gradient
norms of (real) data points ($\left\Vert g_{0}\right\Vert _{2}$ in
Eq.~\ref{eq:x_star_approx}) (Fig.~\ref{fig:iter_approx_grad_norm_curve})
and causes the model to learn faster early in training. However, if
$\alpha$ is too large (10.0), the model performance tends to degrade
over time. If $\alpha$ is too small (0.1), the results are usually
suboptimal when $s$ is small (2) and many update steps are required
to achieve good results (8). The best setting of the GA update is
$(\alpha,s)$=(0.1,8) at which the error is 17.21, smaller than the
error in case the direct approximation is used (17.96). (Results of
the PGA update are presented in the appendix)

\paragraph{How important is finding the most uncertain virtual points?}

We define a ``random regularization'' loss $\Loss_{\text{RR}}$
which has the same formula as $\Loss_{\text{MUR}}$ in Eq.~\ref{eq:MUR_loss}
except that it acts on a \emph{random} virtural points $\hat{x}^{*}$
instead of the most uncertain virtual point $x^{*}$. $\hat{x}^{*}$
is computed as follows:
\[
\hat{x}^{*}=x_{0}+r\times\frac{u}{\left\Vert u\right\Vert _{2}}
\]
where $u$ is a random vector/tensor whose elements are drawn independently
from a standard Gaussian distribution $\Normal(0,1)$. Choosing $u$
like this ensures that $\hat{x}^{*}$ is sampled \emph{uniformly}
on the sphere of radius $r$ centered at $x_{0}$ \cite{muller1959note}.

We compare MT+MUR (with the direct approximation of $x^{*}$) against
MT combined with $\Loss_{\text{RR}}$ (denoted as MT+RR) w.r.t. different
values of $r$ and show the results in Fig.~\ref{fig:MT+RR}. The
advantage of finding the most uncertain virtual points is clear when
$r$ is not too big (e.g., 7 or 10). However, when $r$ becomes bigger
and bigger (e.g., 20 or 40), this advantage disappears and MT+MUR
performs similarly to MT+RR. We think the main reason is that when
$r$ is big, the direct approximation of $x^{*}$ (Eq.~\ref{eq:x_star_approx})
is no longer correct, making $x^{*}$ look more like a random point.

\paragraph{Maximum Uncertainty Training (MUT)}

It is important to know how well MUT performs compared to VAT and
$\Pi$-model. To this end, we train MUT using the same settings for
training $\Pi$-model on the SVHN dataset. The classification errors
of MUT w.r.t. different values of the radius $r$ are shown in Fig.~\ref{fig:MUT_errors}.
At $r=0.5$, MUT with the direct approximation of $x^{*}$ (blue bars
in Fig.~\ref{fig:MUT_errors}) achieves the best error of 3.58$\pm$0.08
which is smaller the reported errors of VAT \cite{miyato2018virtual}
(3.86$\pm$0.11) and $\Pi$-model \cite{laine2016temporal} (4.82$\pm$0.17),
and the error of our own implemented $\Pi$-model (5.18$\pm$0.13
in Table~\ref{tab:SVHN_results}). However, the results become worse
as $r$ increases. We believe the reason is that without the local
smoothness term $\Loss_{\Pi\text{,cons}}(\theta,\theta_{\text{sg}})$
in MUT, the true most uncertain point $x^{*}$ usually lies \emph{closely}
to the real data point $x_{0}$ while the direct approximation $\tilde{x}^{*}=x_{0}+r\frac{g_{0}}{\left\Vert g_{0}\right\Vert _{2}}$
is always $r$ \emph{away from} $x_{0}$. Therefore, if $r$ is not
small enough, $\tilde{x}^{*}$ is no longer a correct approximation
of $x^{*}$, which eliminates the smoothing effect of $\Loss_{\text{MUR}}$.
Luckily, we can relax the distance constraint of $\tilde{x}^{*}$
by using the projected gradient ascent (PGA) update instead. To this
end, the distance between $\tilde{x}^{*}$ and $x_{0}$ can be smaller
than $r$, which gives us more freedom in choosing large $r$. This
can be seen from Fig.~\ref{fig:MUT_errors} as the performance of
MUT with the PGA approximation of $x^{*}$ is almost unaffected by
$r$.

\section{Related Work\label{sec:Related-Work}}

Semi-supervised learning (SSL) is a long-established research area
with diverse approaches \cite{chapelle2006semi}. Within the scope
of this paper, we mainly focus on ``consistency regularization''
(CR) based methods. Details about other methods can be found in \cite{van2019survey}
and \cite{ouali2020overview}. CR based methods aim at learning a
smooth classifier by forcing it to give similar predictions for different
perturbed inputs. There are many ways to define data perturbation
but usually, the more distinct the two perturbations are, the smoother
the classifier is. Standard methods such as Ladder Network \cite{rasmus2015semi},
$\Pi$-model \cite{laine2016temporal} and Mean Teacher \cite{tarvainen2017mean}
perturb data by applying small additive Gaussian noise and binary
dropout to the input and hidden activations. VAT \cite{miyato2018virtual}
and VAdD \cite{park2018adversarial} use adversarial noise and adversarial
dropout for perturbation, respectively. Both can be viewed as performing
\emph{selective smoothing} because they only flatten the input-output
manifold along the direction that gives the largest variance in class
prediction.

Data augmentation is another perturbation technique which is more
effective than general noise injection in specific domains because
it exploits the intrinsic domain structures \cite{xie2019unsupervised}.
For example, augmenting image data with different color filters and
affine transformations encourages the classifier to be insensitive
to changes in color and shape \cite{cubuk2020randaugment}. In case
of text, using thesaurus \cite{zhang2015character,mueller2016siamese}
and back translation \cite{sennrich2016improving,edunov2018understanding}
for augmentation makes the classifier robust to different paraphrases.
These generalization capabilities cannot be obtained with general
noise injections. However, data augmentation also comes with several
limitations such as domain dependence and requirement for external
knowledge not available in the training data.

Apart from data augmentation, standard CR based methods can also be
improved by using additional smoothness-inducing objectives. For example,
SNTG \cite{luo2018smooth} introduces a new loss that forces every
two data points with similar pseudo labels (up to a certain confidence
level) to be close in the low-dimensional feature space. ICT \cite{verma2019interpolation}
is a variant of MT which leverages MixUp \cite{zhang2017mixup} to
encourage linearity of the classifier. Holistic methods like MixMatch
\cite{berthelot2019mixmatch}, ReMixMatch \cite{berthelot2019remixmatch}
and FixMatch \cite{sohn2020fixmatch} combine different advanced techniques
in SSL such as strong data augmentation \cite{cubuk2020randaugment},
MixUp \cite{zhang2017mixup}, entropy minimization \cite{grandvalet2005semi}
and pseudo labeling \cite{lee2013pseudo} into a unified framework
that performs well yet uses very few labeled data.

\section{Conclusion}

We have presented VBI and MUR - two general methods for improving
SSL. We have demonstrated that combining existing CR based methods
with VBI and MUR significantly reduces errors of these methods on
various benchmark datasets. In the future, we would like to incorporate
the likelihood of unlabeled data into VBI to learn a better posterior
distribution of classifier's weights. We also want to apply MUR to
other machine learning problems that demand robustness and generalization.

\newpage\bibliographystyle{aaai21}
\bibliography{bayes_semi}

\newpage{}

\appendix

\section{Appendix}

\subsection{Datasets\label{subsec:Datasets}}

SVHN and CIFAR-10 have 10 output classes while CIFAR-100 has 100 output
classes. All the datasets contain $32\times32$ RGB images as input.
Other details such as the numbers of training and testing samples
for each dataset are provided in Table~\ref{fig:dataset-statistics}.
Following previous works \cite{athiwaratkun2018there,laine2016temporal,tarvainen2017mean},
we preprocessed SVHN by normalizing the input images to zero mean
and unit variance. For CIFAR-10 and CIFAR-100, we applied ZCA whitening
to the input images. After the preprocessing step, we perturbed images
with additive Gaussian noise, random translation, and random horizontal
flip. Details of these transformations are given in Table~\ref{tab:CNN_13_arch}.

\begin{figure}
\begin{centering}
\begin{tabular}{|c|c|c|c|c|}
\hline 
Dataset & image size & \#classes & \#train & \#test\tabularnewline
\hline 
\hline 
SVHN & 32$\times$32 & 10 & 73,257  & 26,032 \tabularnewline
\hline 
CIFAR-10 & 32$\times$32 & 10 & 50,000 & 10,000\tabularnewline
\hline 
CIFAR-100 & 32$\times$32 & 100 & 50,000 & 10,000\tabularnewline
\hline 
\end{tabular}
\par\end{centering}
\caption{Statistics about the datasets used in this paper.\label{fig:dataset-statistics}}
\end{figure}

\subsection{Model Architecture\label{subsec:Model-Settings}}

Following previous works \cite{athiwaratkun2018there,laine2016temporal,tarvainen2017mean,verma2019interpolation},
we use a 13-layer CNN architecture (ConvNet-13) which is shown in
Table~\ref{tab:CNN_13_arch}. Batch Norm \cite{ioffe2015batch} (momentum$=$0.999,
epsilon$=$1e-8) and Leaky ReLU ($\alpha$$=$0.1) are applied to
every convolutional layer of the ConvNet-13.

For models using variational dropout (VD), we simply replace the weight
$w$ of every convolutional/fully-connected layer of the ConvNet-13
with two trainable parameters $\theta$ and $\log\sigma^{2}$. We
also remove all dropout layers since they are unnecessary. Other settings
are kept unchanged.

\begin{table}
\begin{centering}
\begin{tabular}{ll}
Layer & Hyperparameters\tabularnewline
\hline 
Input & 32$\times$32 RGB image\tabularnewline
Translation$^{a}$ & Random $\{\Delta x,\Delta y\}\sim[-4,4]$\tabularnewline
Horizontal flip$^{b}$ & Random $p=0.5$\tabularnewline
Gaussian noise & $\sigma=0.15$\tabularnewline
Convolutional  & 128 filters, $3\times3$, \emph{same} padding\tabularnewline
Convolutional  & 128 filters, $3\times3$, \emph{same} padding\tabularnewline
Convolutional  & 128 filters, $3\times3$, \emph{same} padding\tabularnewline
Max pooling & $2\times2$\tabularnewline
Dropout & $p=0.5$\tabularnewline
Convolutional  & 256 filters, $3\times3$, \emph{same} padding\tabularnewline
Convolutional  & 256 filters, $3\times3$, \emph{same} padding\tabularnewline
Convolutional  & 256 filters, $3\times3$, \emph{same} padding\tabularnewline
Max pooling & $2\times2$\tabularnewline
Dropout & $p=0.5$\tabularnewline
Convolutional & 512 filters, $3\times3$, \emph{valid} padding\tabularnewline
Convolutional & 256 filters, $1\times1$, \emph{valid} padding\tabularnewline
Convolutional & 128 filters, $1\times1$, \emph{valid} padding\tabularnewline
Avg. pooling & $6\times6$ $\rightarrow$ $1\times1$ pixels\tabularnewline
Fully connected & softmax, $128$ $\rightarrow$ $10$ \tabularnewline
\hline 
\end{tabular}
\par\end{centering}
\smallskip{}

\caption{Architecture of the 13-layer CNN used in our paper, similar to those
used in \cite{laine2016temporal,tarvainen2017mean}. The above settings
are for CIFAR-10/CIFAR-100. For SVHN, two small changes are made $^{a}$:
$\{\Delta x,\Delta y\}\sim[-2,2]$ and $^{b}$: no horizontal flip.\label{tab:CNN_13_arch}}
\end{table}

\subsection{Training Settings\label{subsec:Training-Settings}}

Here, we provide details about the settings we use for training our
models. Unless otherwise specified, in all cases, when we say ``ramp
up to $a$'', it means increasing the value from 0 to $a$ during
the \emph{first} $t_{\text{ru}}$ training steps ($t_{\text{ru}}$
is a predefined ramp-up length) using a sigmoid-shaped function $e^{-5(1-x)^{2}}$
where $x\in[0,1]$. On the other hand, when we say ``ramp down from
$a$'', it means decreasing the value from $a$ to 0 during the \emph{last}
$t_{\text{rd}}$ training steps using another sigmoid-shaped function
$1\text{\textminus}e^{-12.5x^{2}}$ where $x\in[0,1]$. The ramp-up
and ramp-down functions are taken from \cite{tarvainen2017mean}.

\paragraph{CIFAR-10 and CIFAR-100}

We apply VD/MUR to MT \cite{tarvainen2017mean} and ICT \cite{verma2019interpolation}
in the experiments on CIFAR-10/CIFAR-100. For both MT and ICT, the
teacher updating momentum is fixed at 0.99. We use a Nesterov momentum
optimizer with the momentum set to 0.9, the L2 weight decay coefficient
set to 0.0001\footnote{For models using VD, we only apply the weight decay to $\theta$,
not to $\log\sigma^{2}$.}, and the learning rate ramped up to and ramped down from 0.1. The
coefficient of the default consistency loss ($\lambda_{1}(t)$ in
Eq.~7, main text) is ramped up to 10\footnote{Previous works use 100 since they take the average over all classes
while we sum over all classes.}. In case VD is used, the coefficient of the KL divergence of weights
($\lambda_{2}(t)$ in Eq.~7, main text) is ramped up to 0.05. In
case MUR is used, the coefficient of the MUR loss ($\lambda_{3}(t)$
in Eq.~7, main text) is ramped up to 4, the radius $r$ is set to
10 if the dataset is CIFAR-10 and 20 if the dataset is CIFAR-100.
The total number of training steps is 280k. The ramp-up and ramp-down
lengths are 10k and 80k steps, respectively. Each batch contains 100
samples with 25 labeled and 75 unlabeled.

\paragraph{SVHN}

We apply VD/MUR to MT \cite{tarvainen2017mean} and $\Pi$-model \cite{laine2016temporal}
in the experiments on SVHN. For MT, the teacher updating momentum
is ramped up from 0.99 to 0.999 using a step function as in \cite{tarvainen2017mean}.
We train both models using a Nesterov momentum optimizer with the
momentum fixed at 0.9. For MT, the L2 weight decay coefficient is
set to 0.0002, the learning rate is ramped up to and ramped down from
0.03, $\lambda_{1}(t)$, $\lambda_{2}(t)$, and $\lambda_{3}(t)$
are ramped up to 12, 0.05, and 2, respectively. On the other hand,
for $\Pi$-model, the L2 weight decay coefficient is set to 0.0001,
the learning rate is ramped up to and ramped down from 0.01, $\lambda_{1}(t)$,
$\lambda_{2}(t)$, and $\lambda_{3}(t)$ are ramped up to 10, 0.05,
and 4, respectively. In case MUR is used, the radius $r$ is set to
10 for both models. The total number of training steps is 280k and
the ramp-up length is 40k for both models. The ramp-down length is
0 for MT and 80k for $\Pi$-model. The batch size is 100 with 25 labeled
and 75 unlabeled for both models.

\subsection{Background on the Local Reparameterization Trick and Variational
Dropout\label{subsec:Background-on-Variational}}

The \emph{local reparameterization trick} \cite{kingma2015variational}
is based on an observation that if $w\in\Real^{M\times N}$ is a factorized
Gaussian random variable with $w_{ij}\sim\Normal(w_{ij};\theta_{ij},\sigma_{ij}^{2})$
($1\le i\le M$, $1\leq j\leq N$) then $z=w^{\top}x$ ($x\in\Real^{M}$,
$z\in\Real^{N}$) is also a factorized Gaussian random variable with
$z_{j}\sim\Normal(z_{i};\nu_{j},\omega_{j}^{2})$ and $\nu_{j}$,
$\omega_{j}^{2}$ are given by:
\[
\nu_{j}=\sum_{i=1}^{M}\theta_{ij}x_{i}\ \ \ \text{and}\ \ \ \omega_{j}^{2}=\sum_{i=1}^{M}\sigma_{ij}^{2}x_{i}^{2}
\]
This trick allows us to convert weight ($w$) sampling into activation
($z$) sampling which is much faster and provides more stable gradient
estimates. Using this trick, Kingma et. al. \cite{kingma2015variational}
further showed that VBI with a factorized Gaussian $q_{\phi}(w)$
was related to Gaussian dropout \cite{wang2013fast} and derived a
new generalization of Gaussian dropout called \emph{variational dropout}
(VD)\emph{.} To see the intuition behind this, we start from the original
formula of Gaussian dropout which is given by:
\begin{equation}
z=\theta^{\top}(x\odot\xi)\label{eq:Gaussian_dropout}
\end{equation}
where $\theta\sim\Real^{M\times N}$ is a deterministic weight matrix;
$\xi\in\Real^{M}$ is a noise vector whose $\xi_{i}$ are sampled
independently from $\Normal(\xi_{i};1,\alpha_{i})$. Inspired by the
local reparameterization trick, we can write Eq.~\ref{eq:Gaussian_dropout}
as $z=w^{\top}x$ where $w\in\Real^{M\times N}$ is a random weight
matrix obtained by perturbing $\theta$ with the multiplicative noise
$\xi$. We can easily show that $w_{ij}$ has a Gaussian distribution
$\Normal(w;\theta_{ij},\alpha_{i}\theta_{ij}^{2})$. In VBI, this
distribution corresponds to the variational posterior distribution
$q_{\phi_{ij}}(w_{ij})$ of an individual weight $w_{ij}$, hence,
Eq.~\ref{eq:Gaussian_dropout} is referred to as \emph{variational
dropout}. The main difference between Gaussian dropout and variational
dropout is that in Gaussian dropout, the dropout rate $\alpha$ is
fixed while in variational dropout, $\alpha$ can be learned\footnote{More precisely, in \cite{kingma2015variational}, $\alpha_{i}$, $\theta_{ij}$
are learnable parameters and $\sigma_{ij}^{2}=\alpha_{i}\theta_{ij}^{2}$
while in \cite{molchanov2017variational}, $\theta_{ij}$, $\sigma_{ij}$
are learnable parameters and $\alpha_{ij}=\frac{\sigma_{ij}^{2}}{\theta_{ij}^{2}}$} to adapt to the data via optimizing the ELBO w.r.t. $w$ (Eq.~6,
main text).

The dropout rate $\alpha$ can also be used to compute the model's
\emph{sparsity}. Based on a theoretical result showing that Gaussian
dropout with $\alpha=\frac{p}{1-p}$ is equivalent to binary dropout
with the dropout rate $p$ \cite{srivastava2014dropout}, we remove
a weight component $w_{ij}$ (by setting its value to 0) if $\log\alpha_{ij}\leq3$
as this corresponds to $p_{ij}\geq0.95$.

\subsection{Why is VD chosen in this work?}

In order to train a large scale Bayesian neural network (BNN) with
the loss in Eq.~7, main text, it is important to choose a VBI method
that supports efficient sampling of the random weights $w$. We choose
VD because it satisfies the above requirement and is flexible enough
to be applied to different network architectures. However, we note
that our proposed method is not limited to VD, and any VBI method
more advanced than VD can be used as a replacement. We discuss some
possible alternatives below, real implementations will be left for
future work.

One promising alternative to train large scale BNNs is sampling-free
VI \cite{haussmann2020sampling,hernandez2015probabilistic,wu2018deterministic}
which leverages moment propagation to derive an analytical expression
for the reconstruction term of the ELBO. To the best of our knowledge,
the most recent advance along this line of works is Deterministic
Variational Inference (DVI) \cite{wu2018deterministic}. However,
this method only supports networks with Heaviside or ReLU activation
functions, and in case of classification, it can only approximate
the reconstruction term (Appdx. B4 in \cite{wu2018deterministic}).

Another interesting approach is leveraging natural gradient with adaptive
weight noise \cite{zhang2018noisy,khan2018fast} to learn the variational
distribution of weights. This approach has several advantages: i)
it can be easily integrated into existing optimizers such as Adam,
and ii) it supports the weight prior with different structures. However,
it is not clear how the natural gradient can be computed for our model.

In VBI, the form of the prior distribution $p(w)$ is also important
as it affects how well the weights can be compressed \cite{louizos2017bayesian,louizos2017learning,neklyudov2017structured}
and how well $q_{\phi}(w)$ can match the true posterior. However,
choosing a good prior is not the main focus of this paper so we simply
choose $p(w)$ to be the log-uniform prior, which is the default prior
for VD.

\subsection{Derivation of the direct approximation of $x^{*}$ in Eq.~10, main
text\label{subsec:Derivation_of_x_star}}

Recall that in Eq.~9, main text, $x^{*}$ is defined as follows:
\begin{equation}
x^{*}=\argmax xH(p(y|x))\ \ \ \text{s.t.}\ \ \ \left\Vert x-x_{0}\right\Vert _{2}\leq r\label{eq:x_star_recall}
\end{equation}
The above objective may have multiple local optima (w.r.t. $x$).
To simplify the problem, we replace $f(x)=H(p(y|x))$ with its linear
approximation derived from the first-order Taylor expansion:
\begin{equation}
f(x)\approx f(x_{0})+{f'(x_{0})}^{\top}(x-x_{0})\label{eq:Ent_FirstOrderTaylor}
\end{equation}
This approximation is usually a lower bound of $H(p(y|x))$ (see the
next section for an explanation). Thus, instead of solving the optimization
problem in Eq.~\ref{eq:x_star_recall}, we solve the following problem:
\begin{equation}
x^{*}\approx\argmax xf(x_{0})+{f'(x_{0})}^{\top}(x-x_{0})\ \ \ \text{s.t.}\ \ \ \left\Vert x-x_{0}\right\Vert _{2}\leq r\label{eq:MU_ConvexOpt}
\end{equation}
With this approximation, the optimization problem in Eq.~\ref{eq:MU_ConvexOpt}
becomes a \emph{convex minimization} problem w.r.t. $x$\footnote{Eq.~\ref{eq:MU_ConvexOpt} is even simpler than a normal convex minimization
problem as it is about finding intersections of a hyperplane with
a sphere centered at $x_{0}$ and having a radius $r$.}, which means a solution, if exists, is unique. Using the method of
Lagrange multipliers, we can rewrite Eq.~\ref{eq:MU_ConvexOpt} as
follows:
\begin{align}
\max_{x,\lambda}\mathcal{O}(x,\lambda)\triangleq\  & f(x_{0})+{f'(x_{0})}^{\top}(x-x_{0})-\nonumber \\
 & \lambda\left(\left\Vert x-x_{0}\right\Vert _{2}-r\right)\ \ \ \text{s.t.}\ \ \ \lambda>0\label{eq:MU_Lagrange}
\end{align}
By setting the derivation of $\mathcal{O}(x,\lambda)$ w.r.t. $x_{i}$
(the $i$-th component of $x$) $\forall i\in\{1,...,D\}$ to $0$,
we have:
\begin{align}
 & \frac{\partial\mathcal{O}}{\partial x_{i}}={f'(x_{0})}_{i}-\lambda\frac{x_{i}-x_{0,i}}{\left\Vert x-x_{0}\right\Vert _{2}}=0\nonumber \\
\Leftrightarrow & x_{i}-x_{0,i}=\frac{{f'(x_{0})}_{i}\left\Vert x-x_{0}\right\Vert _{2}}{\lambda}=\frac{g_{0,i}\left\Vert x-x_{0}\right\Vert _{2}}{\lambda}\label{eq:grad_wrt_xi}
\end{align}
where $g_{0,i}={f'(x_{0})}_{i}$. Similarly, by setting the derivation
of $\mathcal{O}(x,\lambda)$ w.r.t. $\lambda$ to 0, we have:
\begin{align}
 & \frac{\partial\mathcal{O}}{\partial\lambda}=\left\Vert x-x_{0}\right\Vert _{2}-r=0\nonumber \\
\Leftrightarrow & \left\Vert x-x_{0}\right\Vert _{2}=r\nonumber \\
\Leftrightarrow & \sqrt{\sum_{i=1}^{D}\left(x_{i}-x_{0,i}\right)^{2}}=r\label{eq:grad_wrt_lambda}
\end{align}
Substituting $x_{i}-x_{0,i}$ from Eq.~\ref{eq:grad_wrt_xi} into
Eq.~\ref{eq:grad_wrt_lambda}, we have:
\begin{align}
 & \frac{\left\Vert x-x_{0}\right\Vert _{2}}{\lambda}\sqrt{\sum_{i=1}^{D}g_{0,i}^{2}}=r\nonumber \\
\Leftrightarrow & \lambda=\frac{\left\Vert x-x_{0}\right\Vert _{2}\left\Vert g_{0}\right\Vert _{2}}{r}\label{eq:lambda}
\end{align}
We can see that $\lambda>0$ $\forall x\neq x_{0}$. Applying it to
Eq.~\ref{eq:grad_wrt_xi}, we have:
\begin{align*}
x_{i}-x_{0,i} & =\frac{g_{0,i}\left\Vert x-x_{0}\right\Vert _{2}r}{\left\Vert x-x_{0}\right\Vert _{2}\left\Vert g_{0}\right\Vert _{2}}\\
 & =r\frac{g_{0,i}}{\left\Vert g_{0}\right\Vert _{2}}
\end{align*}
which means $x=x_{0}+r\frac{g_{0}}{\left\Vert g_{0}\right\Vert _{2}}$.

\subsection{Lower bound property of the linear approximation of $H(p(y|x))$\label{subsec:Lower-bound-property}}

The second-order Taylor expansion of $f(x)=H(p(y|x))$ is given by:
\[
f(x)\approx f(x_{0})+{f'(x_{0})}^{\top}(x-x_{0})+\frac{1}{2}{(x-x_{0})}^{\top}\mathcal{H}(x_{0})(x-x_{0})
\]
where $\mathcal{H}(x_{0})$ denotes the Hessian matrix of $H(p(y|x))$
at $x_{0}$. During training, $H(p(y|x_{0}))$ is usually close to
0 as the model becomes more certain about the label of $x_{0}$ (or
we can encourage that by penalizing $H(p(y|x_{0}))$ explicitly \cite{grandvalet2005semi}).
Thus, $x_{0}$ tends to be a local minimum of $H(p(y|x))$ and $\mathcal{H}(x_{0})$
is likely to be positive semi-definite. This suggests that $f(x)$
is usually greater than $f(x_{0})+{f'(x_{0})}^{\top}(x-x_{0})$ in
the neighborhood of $x_{0}$.

\subsection{Projected gradient ascent for the original constrained optimization
problem in Eq.~9, main text\label{subsec:ProjectedGradientAscent}}

A common way to solve the constrained optimization problem in Eq.~9,
main text in an iterative manner is using projected gradient ascent.
The approximation of $x^{*}$ at step $t+1$ ($0\leq t\leq T$) is
given by:
\begin{align}
\tilde{x}_{t+1} & =x_{t}+\alpha\frac{\partial H(p(y|x_{t}))}{\partial x_{t}}\label{eq:grad_ascent_step}\\
x_{t+1} & =\argmin{x,\left\Vert x-x_{0}\right\Vert _{2}\leq r}\frac{1}{2}\left\Vert x-\tilde{x}_{t+1}\right\Vert _{2}^{2}\label{eq:proj_step}
\end{align}
where $\alpha\in\Real^{+}$ is the learning rate. Note that the projection
step (Eq.~\ref{eq:proj_step}) is a convex minimization problem.
It achieves the minimum at $\tilde{x}_{t+1}$ if $\left\Vert \tilde{x}_{t+1}-x_{0}\right\Vert _{2}\leq r$
and at $x_{0}+r\frac{\tilde{x}_{t+1}-x_{0}}{\left\Vert \tilde{x}_{t+1}-x_{0}\right\Vert _{2}}$
if $\left\Vert \tilde{x}_{t+1}-x_{0}\right\Vert _{2}>r$. Thus, $x_{t+1}$
in Eq.~\ref{eq:proj_step} can be rewritten as follows:
\[
x_{t+1}=\begin{cases}
\tilde{x}_{t+1} & \text{if}\ \left\Vert \tilde{x}_{t+1}-x_{0}\right\Vert _{2}\leq r\\
x_{0}+r\frac{\tilde{x}_{t+1}-x_{0}}{\left\Vert \tilde{x}_{t+1}-x_{0}\right\Vert _{2}} & \text{otherwise}
\end{cases}
\]

\begin{figure*}
\begin{centering}
\subfloat[MT]{\begin{centering}
\includegraphics[width=0.23\textwidth]{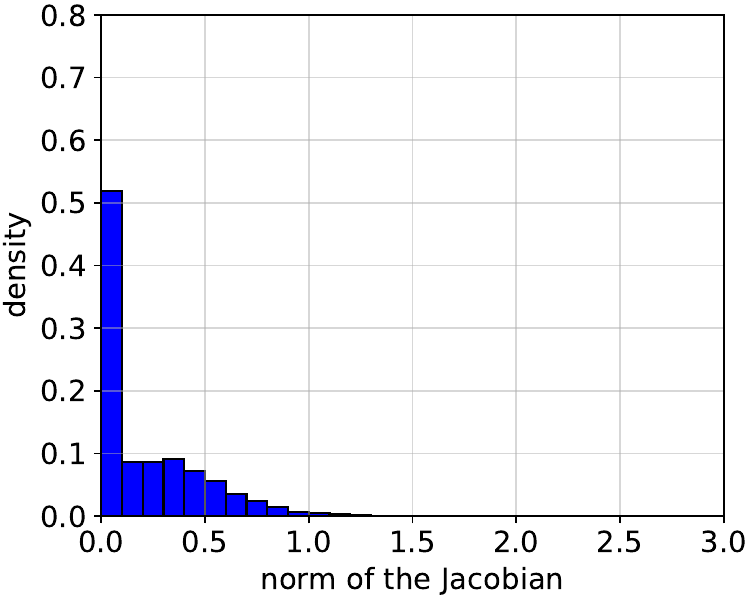}
\par\end{centering}
}\subfloat[MT+VD]{\begin{centering}
\includegraphics[width=0.23\textwidth]{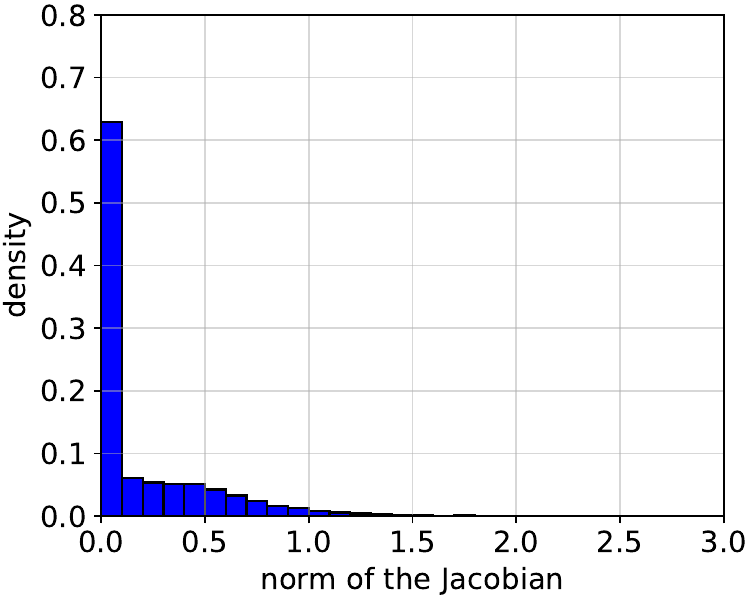}
\par\end{centering}
}\subfloat[MT+MUR]{\begin{centering}
\includegraphics[width=0.23\textwidth]{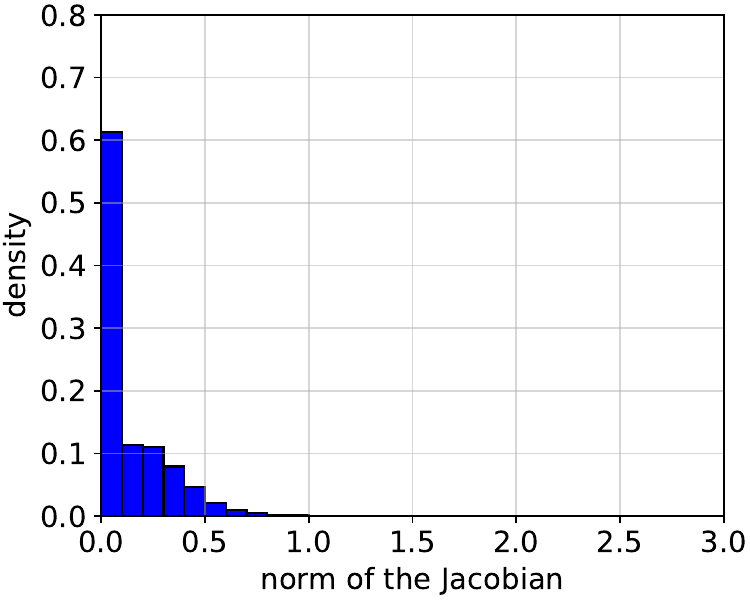}
\par\end{centering}
}\subfloat[MT+VD+MUR]{\begin{centering}
\includegraphics[width=0.23\textwidth]{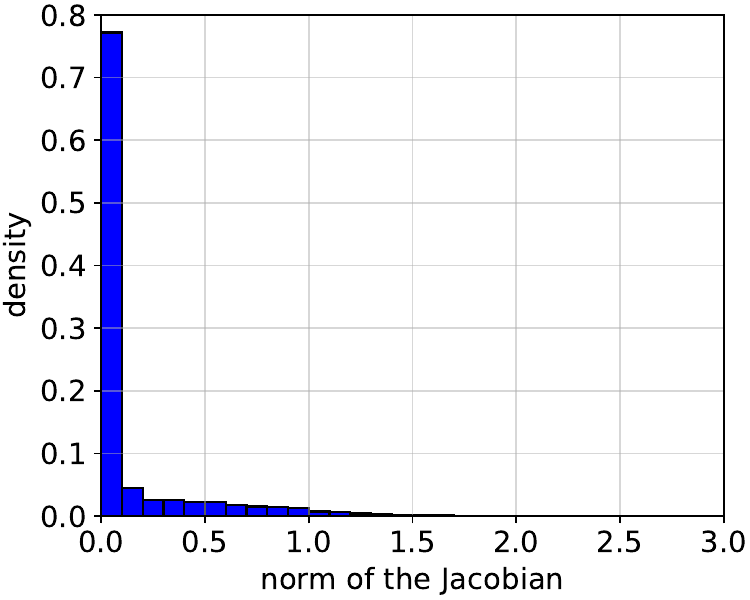}
\par\end{centering}
}
\par\end{centering}
\caption{Sensitivity histograms of MT and its variants trained on CIFAR-10
with 1000 labels.\label{fig:Jacob_norm_hist_MT}}
\end{figure*}

\begin{figure*}
\begin{centering}
\subfloat[ICT]{\begin{centering}
\includegraphics[width=0.23\textwidth]{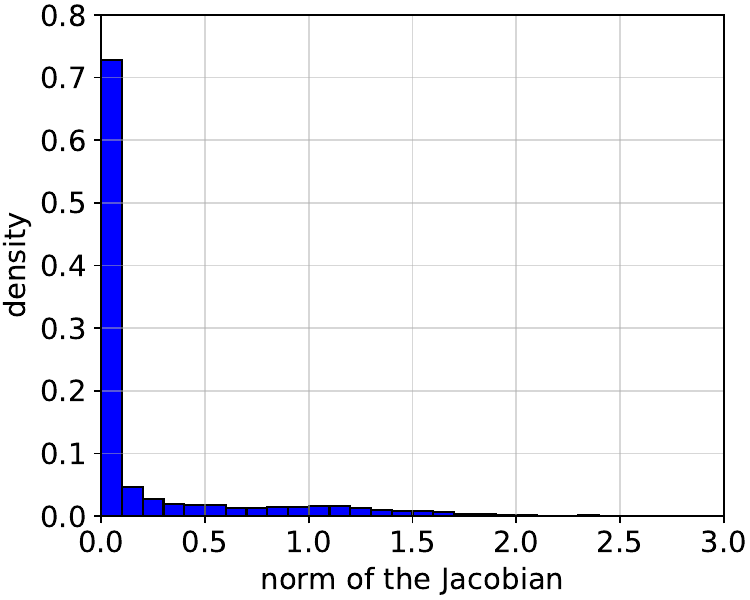}
\par\end{centering}
}\subfloat[ICT+VD]{\begin{centering}
\includegraphics[width=0.23\textwidth]{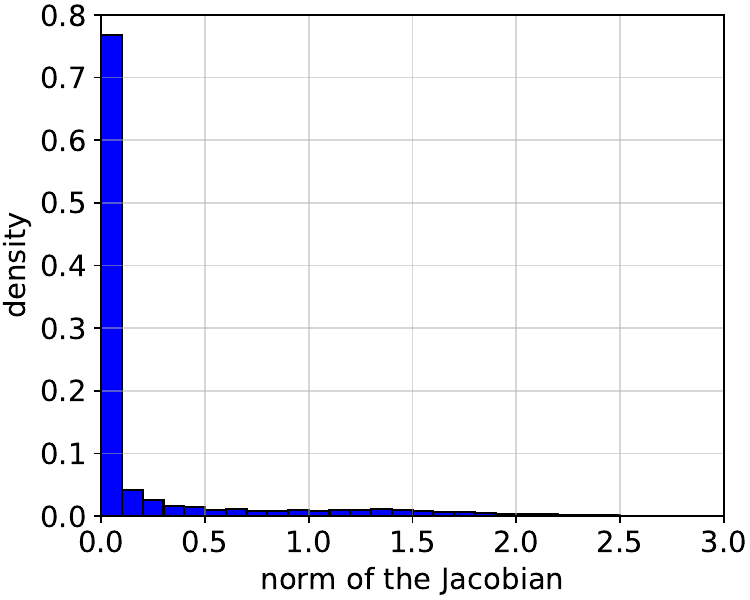}
\par\end{centering}
}\subfloat[ICT+MUR]{\begin{centering}
\includegraphics[width=0.23\textwidth]{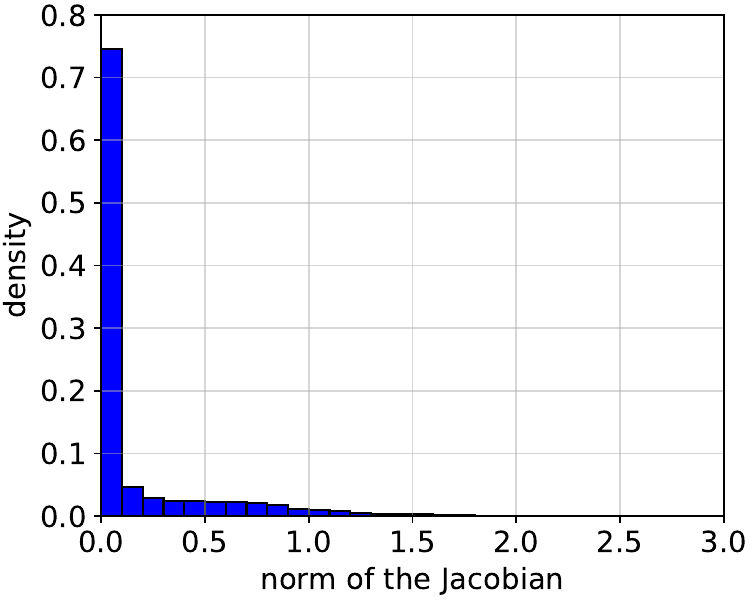}
\par\end{centering}
}\subfloat[ICT+VD+MUR]{\begin{centering}
\includegraphics[width=0.23\textwidth]{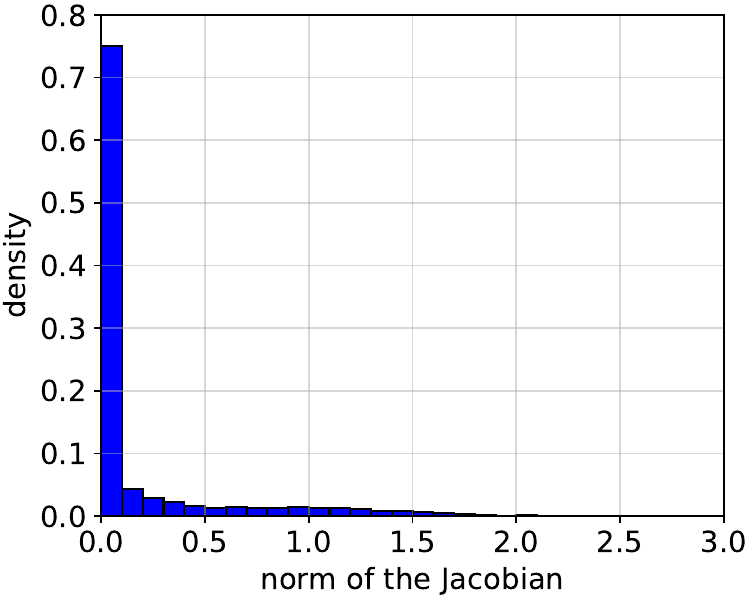}
\par\end{centering}
}
\par\end{centering}
\caption{Sensitivity histograms of ICT and its variants trained on CIFAR-10
with 1000 labels.\label{fig:Jacob_norm_hist_ICT}}
\end{figure*}

\subsection{Gradient ascent for the Lagrangian relaxation problem in Eq.~11,
main text\label{subsec:GradientAscent_LagrangianRelaxation}}

The Lagrangian relaxation of the constrained optimization problem
in Eq.~9, main text is given by:
\begin{equation}
\max_{x,\lambda}\mathcal{F}(x,\lambda)\triangleq f(x)-\lambda\left(\left\Vert x-x_{0}\right\Vert _{2}-r\right)\ \ \ \text{s.t.}\ \ \ \lambda>0\label{eq:Lagrange_4_original}
\end{equation}
where $f(x)=H(p(y|x))$.

One can think of Eq.~\ref{eq:Lagrange_4_original} as a two-step
optimization procedure: First, we maximize $\mathcal{F}(x,\lambda)$
w.r.t. $\lambda$ to find the optimal $\lambda^{*}$ as a function
of $x$. Then, we maximize $\mathcal{F}(x,\lambda^{*})$ w.r.t. $x$
using $\lambda^{*}$ obtained in the previous step. 

However, in the first step, we maximize the approximate of $\mathcal{F}(x,\lambda)$
given in Eq.~\ref{eq:MU_Lagrange} instead of $\mathcal{F}(x,\lambda)$.
This yields $\lambda^{*}(x)$ with similar formula to the one in Eq.~\ref{eq:lambda}:
\begin{equation}
\lambda^{*}(x)=\frac{\left\Vert x-x_{0}\right\Vert _{2}\left\Vert g_{0}\right\Vert _{2}}{r}\label{eq:lambda_star}
\end{equation}
In the second step, we solve the following optimization problem:
\begin{align*}
x^{*}= & \argmax x\mathcal{F}(x)\triangleq f(x)-\frac{\left\Vert x-x_{0}\right\Vert _{2}\left\Vert g_{0}\right\Vert _{2}}{r}\left(\left\Vert x-x_{0}\right\Vert _{2}-r\right)
\end{align*}
 $x^{*}$ can be updated iteratively using vanilla gradient ascent:
\[
x_{t+1}=x_{t}+\alpha\frac{\partial\mathcal{F}(x)}{\partial x}
\]
 with $\frac{\partial\mathcal{F}(x)}{\partial x}$ computed as follows:
\[
\frac{\partial\mathcal{F}(x)}{\partial x}=f'(x)-\frac{\left\Vert g_{0}\right\Vert _{2}(x-x_{0})}{r}\left(2-\frac{r}{\left\Vert x-x_{0}\right\Vert _{2}}\right)
\]

\subsection{Sensitivity histograms of MT, ICT and their variants}

The sensitivity histograms of MT, ICT and their variants are shown
in Figs.~\ref{fig:Jacob_norm_hist_MT}, \ref{fig:Jacob_norm_hist_ICT}.
It is clear that using VD and MUR significantly reduce the classifier's
sensitivity.

\subsection{Visualization of virtual samples\label{subsec:Visualization-of-virtual}}

\begin{figure*}
\begin{centering}
\includegraphics[width=0.98\textwidth]{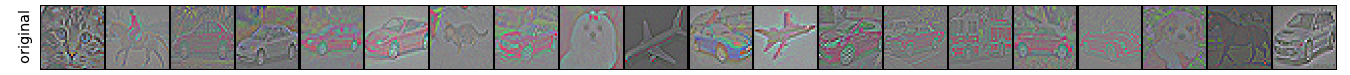}
\par\end{centering}
\begin{centering}
\includegraphics[width=0.98\textwidth]{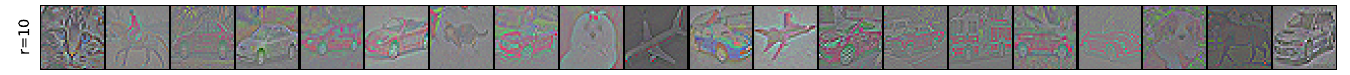}
\par\end{centering}
\begin{centering}
\includegraphics[width=0.98\textwidth]{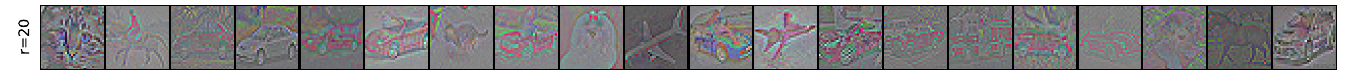}
\par\end{centering}
\begin{centering}
\includegraphics[width=0.98\textwidth]{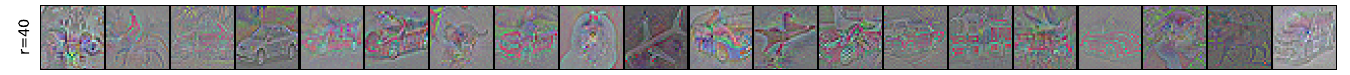}
\par\end{centering}
\caption{Visualization of virtual samples generated by MT+MUR w.r.t. different
radiuses. The model is trained on ZCA whitened CIFAR-10 images.\label{fig:MUR_visualization_ZCA}}
\end{figure*}

To have a better understanding of how virtual samples w.r.t. different
radiuses look like, we show their images in Fig.~\ref{fig:MUR_visualization_ZCA}
after normalizing the pixel values to {[}0, 1{]}. Since the minimum/maximum
pixel values of ZCA whitened CIFAR-10 images are around -30/30, normalization
scales the noise intensity by a factor of 1/60. This is the reason
why we cannot see much difference between the virtual samples at $r=10$
and the input samples.

\begin{figure*}
\begin{centering}
\includegraphics[width=0.98\textwidth]{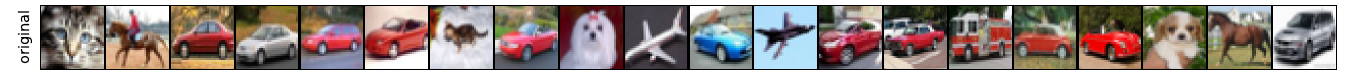}
\par\end{centering}
\begin{centering}
\includegraphics[width=0.98\textwidth]{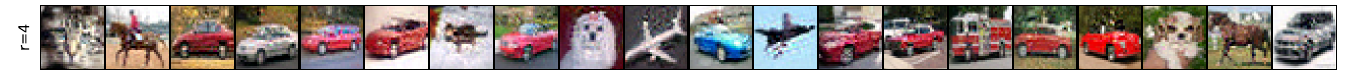}
\par\end{centering}
\begin{centering}
\includegraphics[width=0.98\textwidth]{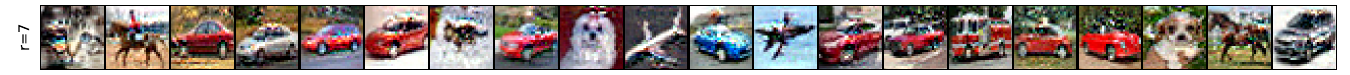}
\par\end{centering}
\begin{centering}
\includegraphics[width=0.98\textwidth]{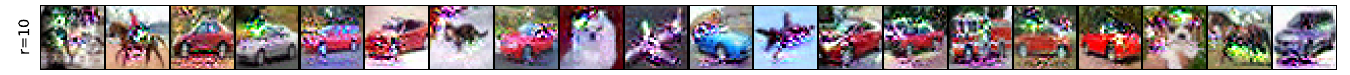}
\par\end{centering}
\begin{centering}
\includegraphics[width=0.98\textwidth]{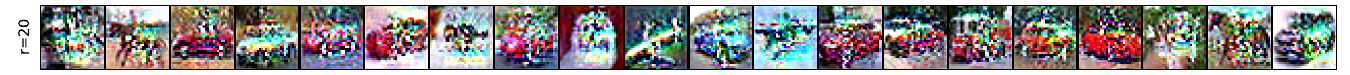}
\par\end{centering}
\begin{centering}
\includegraphics[width=0.98\textwidth]{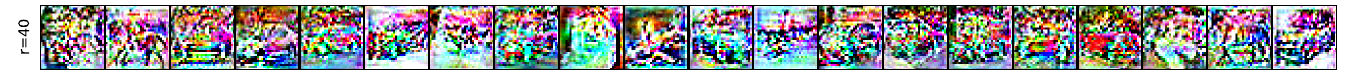}
\par\end{centering}
\caption{Visualization of virtual samples generated by MT+MUR w.r.t. different
radiuses. The model is trained on ZCA whitened CIFAR-10 images.\label{fig:MUR_visualization_trainZCA_testNoZCA}}
\end{figure*}

To improve visualization, we generate virtual samples for unwhitened
images. The perturbed images are clipped in {[}0, 1{]} and are shown
in Fig.~\ref{fig:MUR_visualization_trainZCA_testNoZCA}. It is clear
that when $r$ is small (4, 7), the perturbed images look quite similar
to the input images. By contrast, when $r$ is big (20, 40), most
of the information is lost and the perturb images look very noisy.
A reasonable value of $r$ is 10.

\begin{figure*}
\begin{centering}
\includegraphics[width=0.98\textwidth]{asset/MUR_visualization/x_TrainWithNoZCA}
\par\end{centering}
\begin{centering}
\includegraphics[width=0.98\textwidth]{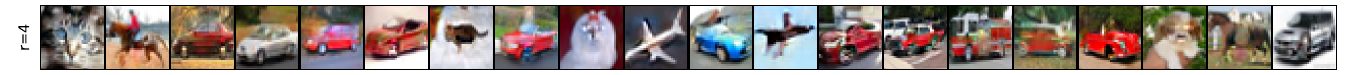}
\par\end{centering}
\begin{centering}
\includegraphics[width=0.98\textwidth]{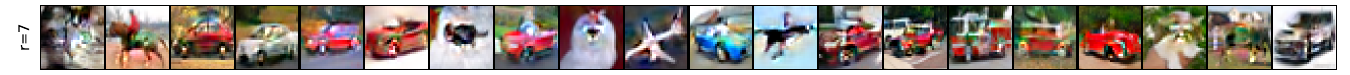}
\par\end{centering}
\begin{centering}
\includegraphics[width=0.98\textwidth]{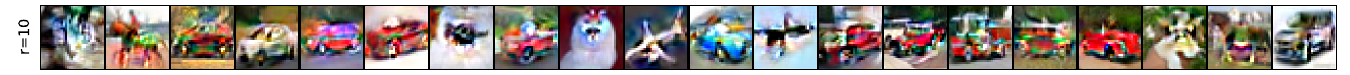}
\par\end{centering}
\begin{centering}
\includegraphics[width=0.98\textwidth]{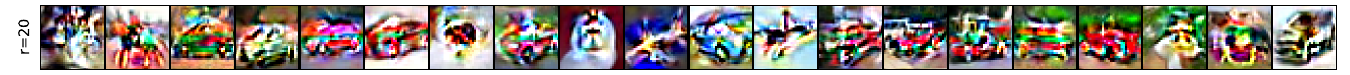}
\par\end{centering}
\begin{centering}
\includegraphics[width=0.98\textwidth]{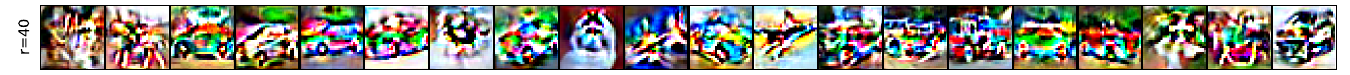}
\par\end{centering}
\caption{Visualization of virtual samples generated by MT+MUR w.r.t. different
radiuses. The model is trained on unwhitened CIFAR-10 images.\label{fig:MUR_visualization_trainNoZCA}}
\end{figure*}

However, since our models are trained on ZCA whitened images, the
visualization in Fig.~\ref{fig:MUR_visualization_trainZCA_testNoZCA}
may not be correct. We address this issue by training a MT+MUR on
unwhitened images. From Fig.~\ref{fig:MUR_visualization_trainNoZCA},
we see that only object features critical for class prediction are
perturbed. Other less important features like background usually remain
unchanged. It suggests that MT+MUR correctly find examples with maximum
uncertainty based on its learned class-intrinsic features.

\begin{figure*}
\begin{centering}
\subfloat[MUR]{\begin{centering}
\includegraphics[height=0.19\textwidth]{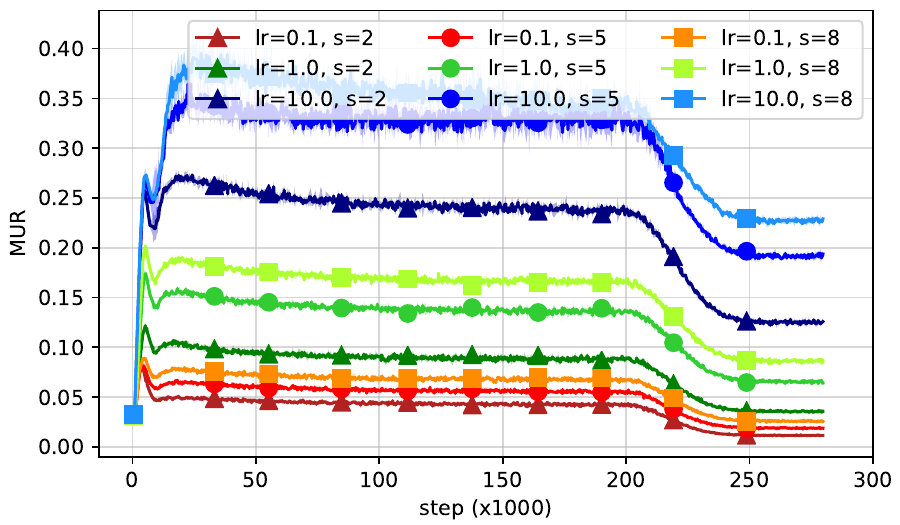}
\par\end{centering}
}\subfloat[consistency loss]{\begin{centering}
\includegraphics[height=0.19\textwidth]{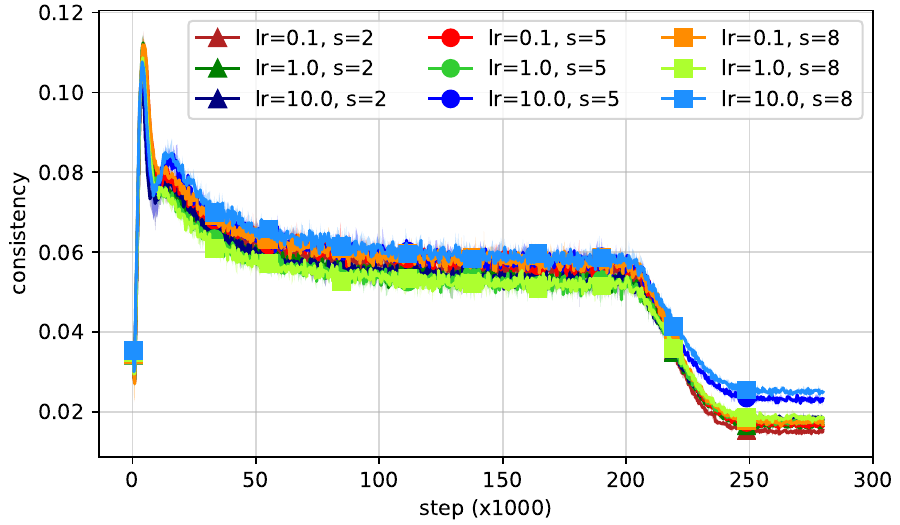}
\par\end{centering}
}\subfloat[cross-entropy loss]{\begin{centering}
\includegraphics[height=0.19\textwidth]{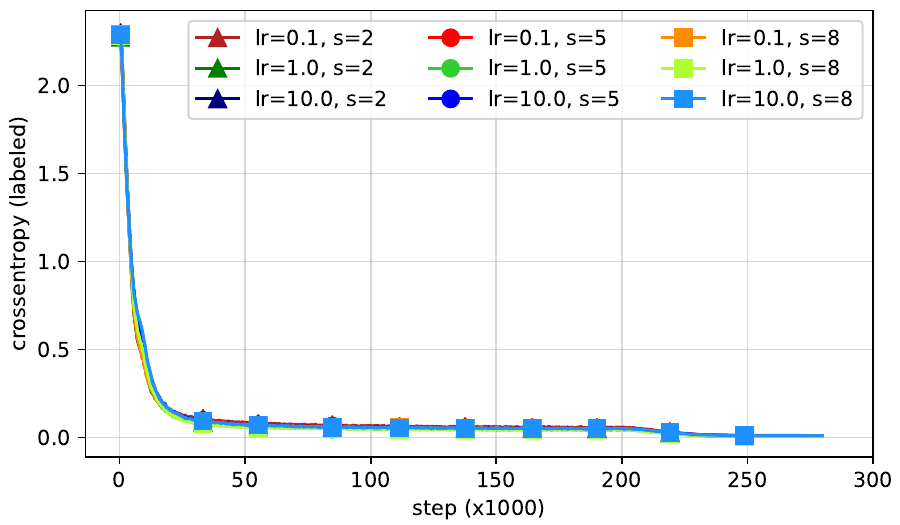}
\par\end{centering}
}
\par\end{centering}
\caption{Loss curves of MT+MUR with the vanilla gradient ascent update of $x^{*}$
w.r.t. different learning rates and numbers of steps. The dataset
is CIFAR-10 with 1000 labels.\label{fig:iter_MT_MUR_GA_losses}}
\end{figure*}

\begin{figure*}
\begin{centering}
\subfloat[classification error]{\begin{centering}
\includegraphics[height=0.2\textwidth]{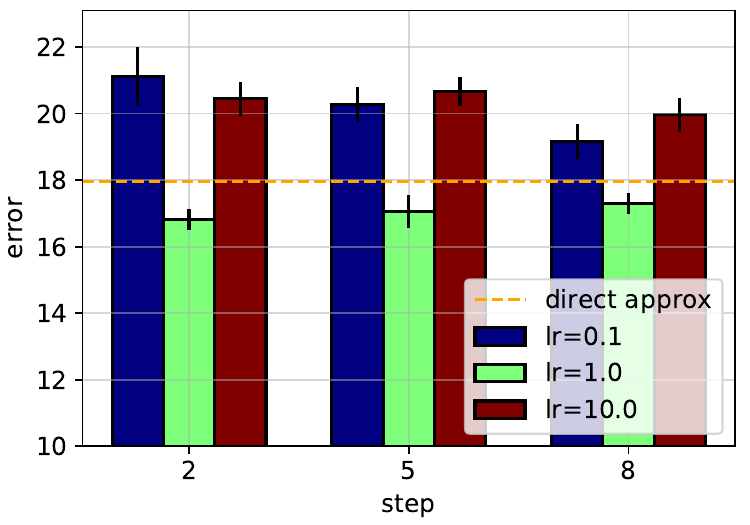}
\par\end{centering}
}\subfloat[$\left\Vert g_{0}\right\Vert $]{\begin{centering}
\includegraphics[height=0.2\textwidth]{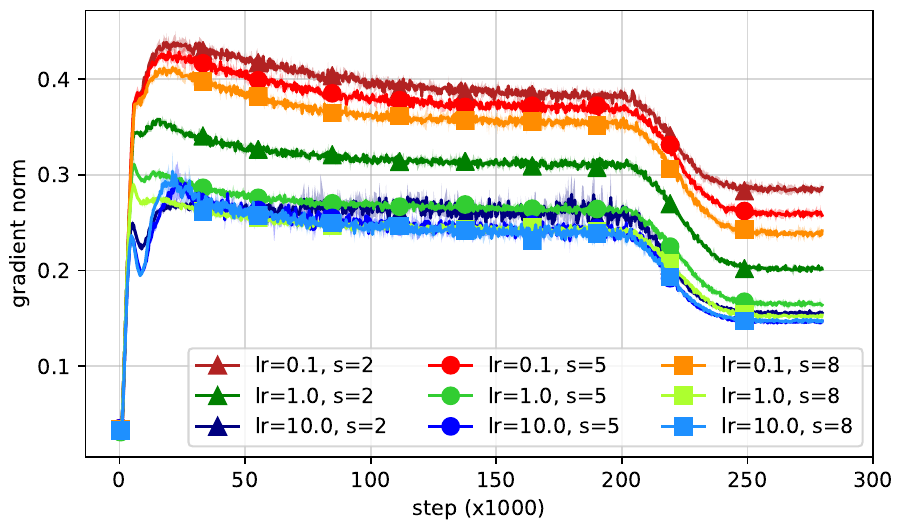}
\par\end{centering}
}\subfloat[MUR]{\begin{centering}
\includegraphics[height=0.2\textwidth]{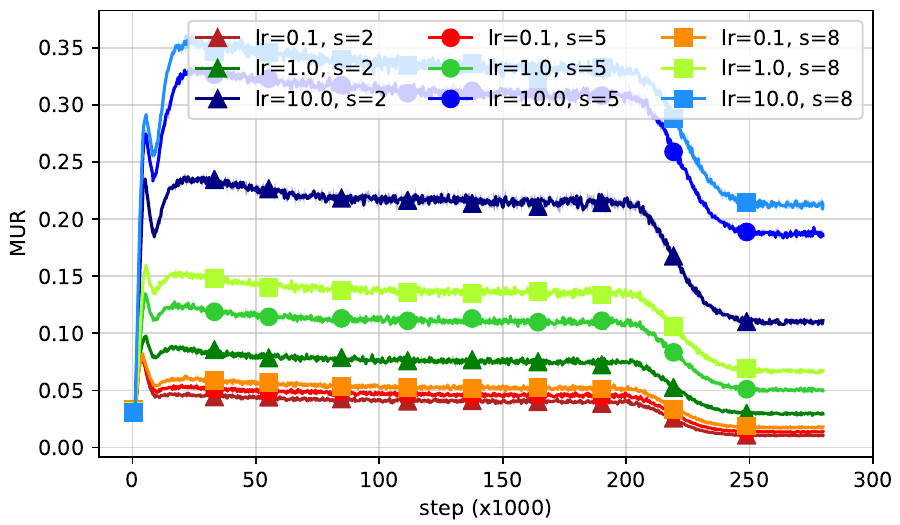}
\par\end{centering}
}
\par\end{centering}
\caption{Results of MT+MUR with the projected gradient ascent update of $x^{*}$
w.r.t. different learning rates and numbers of steps. The dataset
is CIFAR-10 with 1000 labels.\label{fig:iter_MT_MUR_PGA_results}}
\end{figure*}

\subsection{More results of MT+MUR with iterative approximations of $x^{*}$\label{subsec:More-results-iterative-approx}}

Fig.~\ref{fig:iter_MT_MUR_GA_losses} shows the loss curves of MT+MUR
with the vanilla gradient ascent (GA) update of $x^{*}$. While the
MUR loss is affected by different settings of the learning rate $\alpha$
and the number of steps $s$, the consistency and cross-entropy losses
remain almost unchanged. In addition, larger $\alpha$ and larger
$s$ both lead to larger MUR losses since they encourage the model
to find $x^{*}$ farther from $x_{0}$.

\subsubsection{MT+MUR with the projected gradient ascent (PGA) update}

Results of MT+MUR with the projected gradient ascent (PGA) update
of $x^{*}$ are shown in Fig.~\ref{fig:iter_MT_MUR_PGA_results}.
Overall, the dynamics of the PGA update is very similar to that of
the GA update.

\begin{figure}
\begin{centering}
\includegraphics[height=0.19\textwidth]{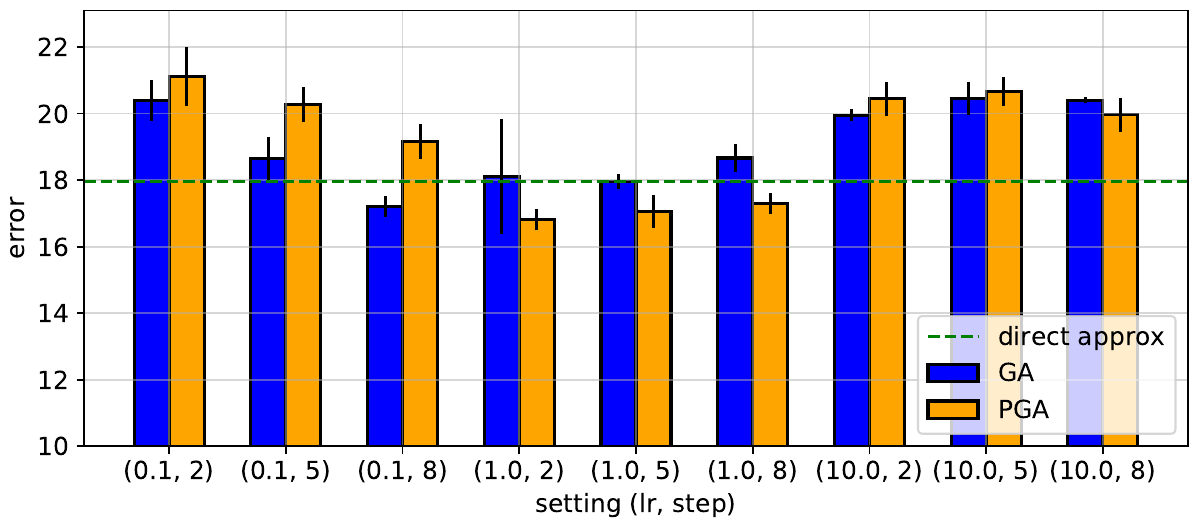}
\par\end{centering}
\caption{Comparison between the GA and PGA updates of $x^{*}$ on different
settings of (learning rate, number of steps).\label{fig:iter_MT_MUR_GA_PGA_compr}}
\end{figure}

We also provide a direct comparison of the classification error between
GA and PGA for every setting of $(\alpha,s)$ in Fig.~\ref{fig:iter_MT_MUR_GA_PGA_compr}.
We see that when $\alpha=0.1$, GA performs better than PGA but when
$\alpha=1.0$, PGA is better. The best result given by PGA is 16.81$\pm$0.31
at $\alpha=1.0$ and $s=2$.

\subsection{Combining VBI and MUR with FixMatch\label{subsec:Combine-with-FixMatch}}

\begin{figure*}
\begin{centering}
\includegraphics[width=0.98\textwidth]{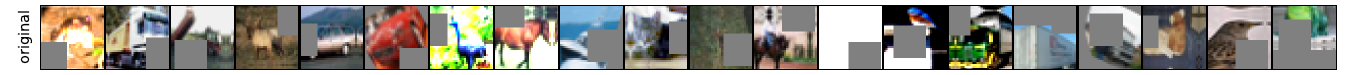}
\par\end{centering}
\begin{centering}
\includegraphics[width=0.98\textwidth]{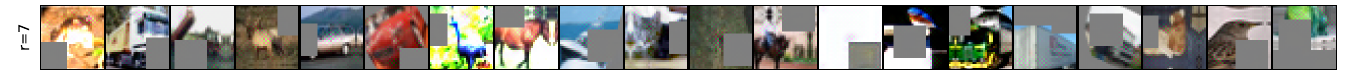}
\par\end{centering}
\begin{centering}
\includegraphics[width=0.98\textwidth]{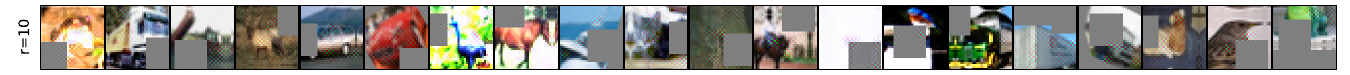}
\par\end{centering}
\caption{Virtual samples generated by FM+MUR. The model is trained on strongly-augmented
CIFAR-10 images. \label{fig:Virtual-samples-FM_MUR}}
\end{figure*}

FixMatch (FM) \cite{sohn2020fixmatch} is currently one of the state-of-the-art
CR based methods that use strong data augmentation \cite{cubuk2019autoaugment,cubuk2020randaugment}.
Its loss function is given by:
\begin{align}
\Loss_{\text{FM}}(\theta)=\  & \Expect_{(x_{l},y_{l})\sim\Data_{l}}\left[-\log p(y_{l}|x_{l},\theta)\right]+\nonumber \\
 & \lambda\Expect_{x_{u}\sim\Data_{u}}\left[-1_{p(\tilde{y}|x_{u},\theta)>\tau}\log p(\tilde{y}|\mathcal{A}(x_{u}),\theta)\right]\label{eq:FM_loss}
\end{align}
where $\mathcal{A}(x_{u})$ denotes the strongly-augmented version
of $x_{\ensuremath{u}}$; $\tilde{y}$ is the pseudo label of $x_{u}$
which satisfies that $\tilde{y}=\argmax kp(k|x_{u})$; $\tau$ is
a confidence threshold.

Since the loss function of FixMatch in Eq.~\ref{eq:FM_loss} also
has the form $\Loss_{\text{xent,}l}(\theta)+\lambda(t)\Loss_{\text{cons}}(\theta,\cdot)$,
we can integrate VBI and MUR into this model. The loss function of
FM+VBI is:
\begin{align*}
 & \Loss_{\text{FM+VBI}}(\phi)=\\
 & \Expect_{w\sim q_{\phi}(w)}\left[\Expect_{(x_{l},y_{l})\sim\Data_{l}}\left[-\log p(y_{l}|x_{l},w)\right]\right]+\\
 & \lambda_{1}\Expect_{w\sim q_{\phi}(w)}\left[\Expect_{x_{u}\sim\Data_{u}}\left[-1_{p(\tilde{y}|x_{u},\theta)>\tau}\log p(\tilde{y}|\mathcal{A}(x_{u}),w)\right]\right]+\\
 & \lambda_{2}D_{KL}(q_{\phi}(w)\|p(w))
\end{align*}
with a note that in the indicator function, we use the mean network
$f_{\theta}$ to compute $p(\tilde{y}|x_{u},\theta)$.

And the loss function of FM+MUR is:
\begin{align*}
\Loss_{\text{FM+MUR}}(\phi)=\  & \Expect_{(x_{l},y_{l})\sim\Data_{l}}\left[-\log p(y_{l}|x_{l},w)\right]+\\
 & \lambda\Expect_{x_{u}\sim\Data_{u}}\left[-1_{p(\tilde{y}|x_{u},\theta)>\tau}\log p(\tilde{y}|\mathcal{A}^{*}(x_{u}),w)\right]
\end{align*}
where $\mathcal{A^{*}}(x_{u})$ is the MUR virtual point of $\mathcal{A}(x_{u})$
found by optimizing the following objective:
\begin{equation}
\mathcal{A}^{*}(x_{u})=\argmax xH(p(y|x))\ \ \ \text{s.t.}\ \ \ \|x-\mathcal{A}(x_{u})\|\leq r\label{eq:MUR_strong_aug}
\end{equation}

In our experiments, we observed that FM+VD gives the same results
as FM, which suggests that weight perturbation has almost no contribution
on generalization when strong data augmentation is available. FM+MUR,
however, is much worse than FM. For example, on CIFAR-10 with 250
labels, FM achieves an average error of 5.07 while FM+MUR can only
achieve an average error of 13.45. We guess the problem is that the
class predictions for strongly-augmented data is much more inconsistent
and inaccurate compared to the class predictions for normal data.
Thus, the MUR optimization problem in Eq.~\ref{eq:MUR_strong_aug}
is often ill-posed and may not return desirable results. We can check
that by visualizing the virtual samples for strongly-augmented data
in Fig.~\ref{fig:Virtual-samples-FM_MUR}.

\end{document}